\documentclass[journal,twoside]{IEEEtran}
\IEEEoverridecommandlockouts

\usepackage{graphicx} \graphicspath{ {figures/} }
\usepackage{amsmath,amssymb,mathabx,amsbsy,mathtools,etoolbox}
\usepackage[utf8]{inputenc} % allow utf-8 input
\usepackage[T1]{fontenc}    % use 8-bit T1 fonts

\usepackage{algorithmic}
\usepackage[ruled]{algorithm2e}
\usepackage{acronym}
\usepackage{enumitem}
\usepackage[breaklinks,colorlinks,linkcolor=black]{hyperref}
\usepackage{balance}
\usepackage{xspace}
\usepackage{setspace}
\usepackage[skip=3pt,font=small]{subcaption}
\usepackage[skip=3pt,font=small]{caption}
\usepackage[dvipsnames,svgnames,x11names]{xcolor}
\usepackage[capitalise,nameinlink]{cleveref}
\usepackage{booktabs,tabularx,colortbl,multirow,multicol,array,makecell}
\usepackage{pifont}
\usepackage{cite}
\usepackage{nicematrix}
\usepackage{dsfont}
\usepackage{dblfloatfix}
\usepackage{url} 
\makeatletter
\DeclareRobustCommand\onedot{\futurelet\@let@token\@onedot}
\def\@onedot{\ifx\@let@token.\else.\null\fi\xspace}
\def\eg{\emph{e.g}\onedot}

\def\etc{\emph{etc}\onedot}

\makeatother

\makeatletter
\def\BState{\State\hskip-\ALG@thistlm}
\makeatother

\makeatletter
\renewcommand{\paragraph}{%
  \@startsection{paragraph}{4}%
  {\z@}{0ex \@plus 0ex \@minus 0ex}{-1em}%
  {\hskip\parindent\normalfont\normalsize\bfseries}%
}
\makeatother

\crefname{algorithm}{Alg.}{Algs.}
\Crefname{algocf}{Algorithm}{Algorithms}
\crefname{section}{Sec.}{Secs.}
\Crefname{section}{Section}{Sections}
\crefname{table}{Tab.}{Tabs.}
\Crefname{table}{Table}{Tables}
\crefname{figure}{Fig.}{Fig.}
\Crefname{figure}{Figure}{Figure}
\definecolor{gblue}{HTML}{4285F4}
\definecolor{gred}{HTML}{DB4437}
\definecolor{ggreen}{HTML}{0F9D58}

\definecolor{mygray}{gray}{.92}

\acrodef{qp}[QP]{Quadratic Programming}
\acrodef{dof}[DoF]{Degree of Freedom}
\acrodef{ros}[ROS]{Robot Operating System}
\acrodef{dof}[DoF]{Degree of Freedom}
\acrodef{com}[CoM]{center of mass}
\acrodef{rl}[RL]{reinforcement learning}
\acrodef{mpc}[MPC]{model predictive control}
\acrodef{com}[CoM]{center of mass}
\acrodef{cam}[CAM]{centroidal angular momentum}
\acrodef{cmm}[CMM]{centroidal momentum matrix} 
\acrodef{cdm}[CDM]{centroidal dynamics model}
\acrodef{cop}[CoP]{Center of Pressure}
\acrodef{srbm}[SRBM]{single rigid body model}
\acrodef{lipm}[LIPM]{linear inverted pendulum model}
\acrodef{slipm}[SLIPM]{spring-loaded inverted pendulum model}
\acrodef{ppo}[PPO]{Proximal Policy Optimization}
\acrodef{to}[TO]{Trajectory Optimization}
\acrodef{cot}[CoT]{cost of transport}
\acrodef{cmdp}[CMDP]{constrained Markov decision process}

\title{ECO: Energy-Constrained Optimization with Reinforcement Learning for Humanoid Walking}

\author{Weidong Huang, Jingwen Zhang,~\IEEEmembership{Member,~IEEE},  Jiongye Li, Shibowen Zhang, Jiayang Wu, \\Jiayi Wang, Hangxin Liu,~\IEEEmembership{Member,~IEEE}, Yaodong Yang,~\IEEEmembership{Member,~IEEE}, Yao Su,~\IEEEmembership{Member,~IEEE}
 \thanks{This work was supported in part by the National Natural Science Foundation of China (No. 62403064, 62403063) and Shenzhen Science and Technology Program (No. ZDCY20250901094531003).
\textit{(Weidong Huang, Jingwen Zhang contributed equally to this work.)} \textit{(Corresponding authors: Jingwen Zhang and Yao Su.)}}
\thanks{Weidong Huang, Jingwen Zhang, Jiongye Li, Shibowen Zhang, Jiayang Wu, Jiayi Wang, Hangxin Liu, Yao Su are with State Key Laboratory of General Artificial Intelligence, Beijing Institute for General Artificial Intelligence (BIGAI), Beijing 100080, China (e-mails: bigeasthuang@gmail.com; zhangjingwen@bigai.ai; lijiongye@bigai.ai; zhangshibowen@bigai.ai; wujiayang@bigai.ai; wangjiayi@bigai.ai; liuhx@bigai.ai; suyao@bigai.ai).}
 \thanks{Jiongye Li is also with Department of Automation, Tsinghua University, Beijing 100084, China.}
 \thanks{Shibowen Zhang is also with Department of Automation, University of Science and Technology of China, Hefei 230022, China.}
\thanks{Jiayang Wu is also with Department of Computer Science, Harbin Institute of Technology, Harbin 150001, China.}
 \thanks{Yaodong Yang is with Institute for Artificial Intelligence and School of Artificial Intelligence, Peking University, Beijing 100871, China (e-mail: yaodong.yang@pku.edu.cn).}}

\begin{document}

\maketitle

\begin{abstract}
Achieving stable and energy-efficient locomotion is essential for humanoid robots to operate continuously in real-world applications. Existing \ac{mpc} and \ac{rl} approaches often rely on energy-related metrics embedded within a multi-objective optimization framework, which require extensive hyperparameter tuning and often result in suboptimal policies. To address these challenges, we propose ECO (\underline{E}nergy-\underline{C}onstrained \underline{O}ptimization), a constrained \ac{rl} framework that separates energy-related metrics from rewards, reformulating them as explicit inequality constraints. This method provides a clear and interpretable physical representation of energy costs, enabling more efficient and intuitive hyperparameter tuning for improved energy efficiency. 
ECO introduces dedicated constraints for energy consumption and reference motion, enforced by the Lagrangian method, to achieve stable, symmetric, and energy-efficient walking for humanoid robots. We evaluated ECO against MPC, standard RL with reward shaping, and four state-of-the-art constrained RL methods. Experiments, including sim-to-sim and sim-to-real transfers on the kid-sized humanoid robot BRUCE, demonstrate that ECO significantly reduces energy consumption compared to baselines while maintaining robust walking performance. These results highlight a substantial advancement in energy-efficient humanoid locomotion. All experimental demonstrations can be found on the project website: \url{https://sites.google.com/view/eco-humanoid}.
\end{abstract}

\def\abstractname{Note to Practitioners}
\begin{abstract}
Traditional \ac{mpc} and \ac{rl} approaches often require extensive hyperparameter tuning and frequently result in suboptimal solutions for improving energy efficiency while maintaining stable walking performance. ECO is designed to address these challenges by reformulating energy consumption as explicit inequality constraints, providing a physically interpretable and intuitive approach to optimizing energy efficiency. This framework is particularly well-suited for applications prioritizing energy conservation and operational stability, such as surveillance, disaster response, and long-duration autonomous operations. Additionally, ECO generates emergent behaviors such as lighter steps and reduced body shaking, which are especially advantageous for loco-manipulation tasks by minimizing disruptions to upper-body manipulation caused by locomotion. Comparative experiments empirically offer valuable insights into constraint selection and learning setups, which may also inspire relevant ongoing research in constrained \ac{rl}.
\end{abstract}


\begin{figure}[t!]
\centering
\includegraphics[width=\linewidth,trim=0cm 7.5cm 10cm 0cm,clip]{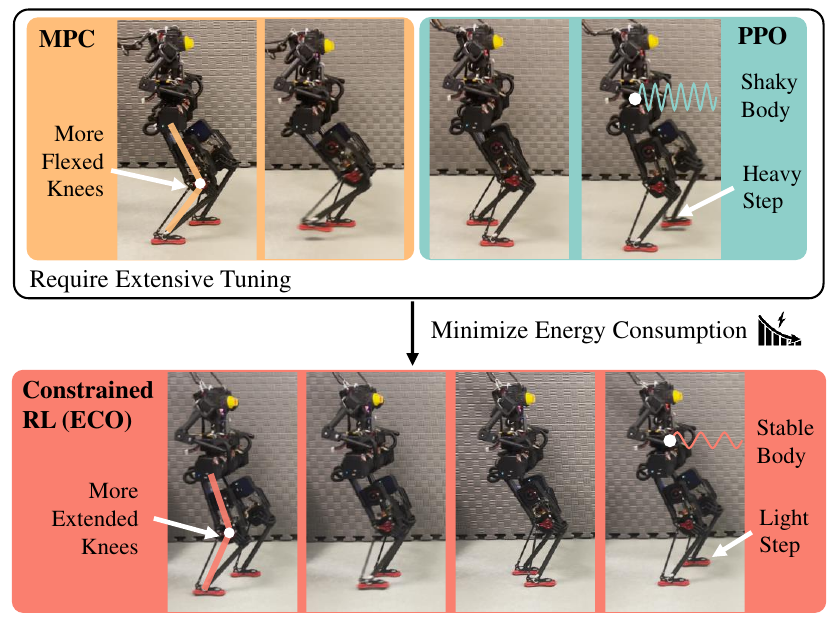}
\caption{\textbf{Comparison between the proposed constrained RL framework, ECO, with MPC and normal RL (PPO) baselines.} It creates synergy between energy consumption and walking stability of the humanoid robots without requiring an extensive parameter-turning process, outperforming both MPC and normal RL baselines.}
\label{fig:moti}
\end{figure}

\section{Introduction}

\IEEEPARstart{R}{ecent} advances in humanoid robotics have enabled these robots to perform complex motions, such as walking, running, jumping, and even loco-manipulation~\cite{wang2024arm,he2024cdm,zhang2023design,liu2025pr2}. Despite these achievements, energy efficiency still remains a significant bottleneck for humanoid robots. Compared to their biological counterparts~\cite{wieber2016modeling}, humanoid robots require substantially more energy to perform similar tasks, limiting their operational range, endurance, and maximum payload in real-world applications~\cite{alqaham2024energy,wieber2016modeling}.

To improve the energy efficiency of humanoid robots, one promising approach is to leverage \acf{rl} to optimize the control policy. For instance, \ac{ppo}-based frameworks incorporate energy-related terms into reward functions, such as minimizing joint torques, joint accelerations, and contact forces~\cite{fu2021minimizing,fu2024humanplus,gu2024humanoid}. 
However, to achieve efficient and stable locomotion, these methods often require extensive tuning on the weights of the reward terms regarding energy cost, task completion, and stability. This tuning process is non-intuitive and time-consuming, as (i) the effects of weight configurations are not straightforward, (ii) the training periods—ranging from hours to days—to evaluate a single coefficient choice even with parallel computation~\cite{jeon2023benchmarking,  chow2018risk}. Furthermore, conflicting objectives within the reward terms can result in suboptimal solutions or even convergence failures~\cite{he2023autocost,huang2023safe}. For example, the learned policy might prioritize minimizing energy consumption at the cost of stability, resulting in unstable walking, or emphasize stability at the expense of significantly increased energy usage. 

To simplify the reward tuning process for optimizing energetically efficient locomotion with \ac{rl}, we propose ECO (\underline{E}nergy-\underline{C}onstrained \underline{O}ptimization). ECO leverages constrained \ac{rl} formualtion to separate conflicting objectives into distinct rewards and constraints~\cite{achiam2017constrained, wang2024arm, kim2024not}. This separation ensures that critical constraints, such as energy consumption limits, are strictly enforced, while task-related rewards, such as velocity tracking and stability can be optimized without extensive tuning of coefficients. Energy constraint thresholds are identified through a physically intuitive tuning process, incrementally adjusted based on direct energy criteria using a linear search algorithm. We evaluate our approach by comparing four state-of-the-art constrained \ac{rl} methods and investigating various constraint settings. Our results empirically show that the proposed setup outperforms others in training stability and constraint enforcement. Through extensive experiments on both simulation and hardware platforms, we validate that ECO consistently converges to solutions that minimize energy consumption while ensuring stable and robust walking.

Our main contributions can be summarized as follows:
\begin{enumerate}
    \item We propose ECO, a method designed for optimizing the energy efficiency of humanoid locomotion. 
    %The core idea is to formulate the problem as a constrained \ac{rl} problem, where the objective is to find a stable locomotion policy that remains with a desired energy consumption limit. 
    Rather than balancing energy minimization with other reward terms—such as velocity tracking, motion tracking, and stability—we explicitly treat energy cost as a constraint. This approach provides a clear, interpretable physical meaning of the energy cost, enabling efficient and intuitive hyperparameter tuning focused on energy efficiency.
        
    \item We investigate four constrained \ac{rl} algorithms and also different constraint settings to train the policy. Through extensive simulations, we empirically find that the PPO-Lagrangian method with the current constraint setup achieves the lowest energy cost with fast and stable convergence. Additionally, instead of manually designing efficient walking, we observe that the resulting policy promotes i) extended knee movements, ii) lighter steps, and iii) reduced body shaking as shown in~\cref{fig:moti}.

    \item We carry out real-world validations on the Bruce robot. To the best of our knowledge, this is the first work to achieve energy-efficient humanoid walking using constrained RL on a real humanoid robot. Compared to MPC and PPO baselines, ECO demonstrates approximately 6 times lower energy consumption than MPC and 2.3 times lower than PPO, setting a new benchmark for energy-efficient humanoid locomotion.
\end{enumerate}

We organize the remainder of the paper as follows. \cref{sec:related} reviews the literature related to this work. \cref{sec:problem} introduces the preliminaries of \ac{rl} and constrinted \ac{rl} formulations. \cref{sec:framework} presents the proposed ECO framework in detail. \cref{sec:sim,sec:exp} show the simulation and experiment result of ECO on BRUCE with comprehensive evaluations and compares with baseline methods. Finally, we discuss and conclude the paper in \cref{sec:discussion,sec:conclusion}.

\section{Related Work}
\label{sec:related}
\subsection{Energy Optimization in Legged Locomotion}
Improving energy efficiency directly helps to extend the endurance of legged robots and prevent actuator overheating, thereby improving the overall performance~\cite{alqaham2024energy}. For example, passive dynamic walkers can achieve efficient and stable walking through mechanical designs without any actuation~\cite{collins2005efficient} or with minimal actuation~\cite{tedrake2004actuating}.
% Efficient but passive dynamic walking has been achieved without any actuation~\cite{collins2005efficient} or with minimal actuation~\cite{tedrake2004actuating}.} 
However, achieving similar efficiency in more complex legged systems, such as quadrupedal and bipedal robots, remains a challenge. As a unified \ac{to} framework has been widely used for legged locomotion~\cite{kuindersma2016optimization, shen2022implementation}, incorporating energy-related constraints tends to be a potential solution. \cite{xi2016selecting} formulates the \ac{to} problem with the \ac{cot} as a constraint, which quantifies the energy expenditure per unit distance. However, using \ac{to} to consider complex constraints often results in a high-dimensional and nonconvex landscape, which hinders real-time usage. To simplify the problem, energy constraints are approximated using the norm of joint torques and accelerations~\cite{ponton2021efficient, he2024cdm,wang2024online}, or by encouraging specific bioinspired efficient behaviors, such as adaptive \ac{com} height control and rolling contact of the stance leg in gait design~\cite{fasano2024efficient}. In these methods, energy consumption is only implicitly considered, which often makes it compromised in favor of other task requirements, such as stability and robustness.

\subsection{Learning for Legged Locomotion}
Given the aforementioned issues, offline learning a locomotion policy with complex rewards can be a promising solution to balance energy-efficiency and robustness in real-time~\cite{cuiadapting, li2023interactive, radosavovic2024humanoid}.\cite{siekmann2021sim} demonstrated sim-to-real learning of multiple bipedal gaits through periodic reward composition. Similarly, \cite{kumar2022adapting} extended rapid motor adaptation (RMA) for bipedal locomotion to address challenges such as payloads and slippery surfaces. Other methods also integrated model-based footstep planning with model-free \ac{rl} to achieve dynamic and efficient walking~\cite{lee2024integrating, bang2024rl}. However, to avoid disrupting the primary learning goals, energy penalties in the reward function often act as regularizers, with minimal impact on reducing excessive joint torque usage. \cite{yu2018learning} balanced energy consumption with stability through curriculum learning with symmetric mirror loss. \cite{10802439} further studied different ways to incorporate symmertry, highlighting how symmetry priors can improve gait robustness and sample efficiency without overly restricting policy exploration. \cite{fu2021minimizing} explored efficient quadrupedal gaits with energy-centric rewards.~\cite{peng2022ase, peng2018deepmimic, xu2022preference, li2024human} train control policies for humanoid locomotion using human motion data as reference trajectories. Despite these advancements, the energy-related hyperparameters, like the reward weight or curriculum length, still present an unclear physical meaning, which intricates the weight tuning and curriculum design.

\subsection{Constrained Reinforcement Learning}
Constrained RL, or safe RL, formulates the problem as a \ac{cmdp}~\cite{cmdp1999, li2024evolutionary, wang2024constrained, non_zero}, using cost functions to decouple rewards from constraints. This allows robots to maximize rewards while keeping costs within physically meaningful thresholds. CPO~\cite{cpo2017} uses second-order optimization to enforce constraints iteratively but struggles with scalability in multi-constraint settings. PPO-Lagrangian (PPO-Lag)~\cite{ray2019benchmarking} enhances PPO with primal-dual updates for better constraint satisfaction and performance. Constraint manifold is defined for safe exploration leading to actions that satisfy corresponding constraints in~\cite{liu2022robot, kicki2023fast}. However, efficient constraint manifold computation for legged robots with a floating base remains challenging. Constraint-rectified Policy Optimization (CRPO)~\cite{xu2021crpo} offers a primal-based framework well-suited for multi-constraint problems, while Interior-point Policy Optimization (IPO)~\cite{liu2020ipo} employs log-barrier penalties for feasibility enforcement. \cite{kim2024not} extended IPO with a teacher-student framework, transferring most reward terms into constraints to avoid reward scale tuning, and demonstrated robust performance on legged robots. Penalized Proximal Policy Optimization (P3O)~\cite{zhang2022penalized} uses clamping penalty functions for constraint enforcement and has shown success in reducing violations (\eg, self-collision, joint speed limits) in quadruped locomotion~\cite{lee2024exploring}. Despite these advancements, these methods have only been validated on quadruped robots. Applying them to humanoid robots, especially when combining energy consumption with other constraints (\eg, self-collision), can overly restrict the search space. Given the smaller convergence region of humanoid robots compared to quadrupeds, rethinking how to select and design constraints and how to select an appropriate optimization method within the constrained RL framework to generate efficient and stable walking performance of humanoid robots remains a critical challenge. 

\section{PRELIMINARIES \& PROBLEM STATEMENT}  \label{sec:problem}

\subsection{Markov Decision Processes}
\ac{rl} is typically modeled as a Markov Decision Process (MDP)~\cite{cmdp1999}: 
\begin{equation}
    \mathcal{M} = (\mathcal{S}, \mathcal{A}, \mathbb{P}, R, \mu, \gamma),
\end{equation}
where $\mathcal{S}$ and $\mathcal{A}$ denote the state and action spaces, respectively. The functions $\mathbb{P}(\pmb{s}'|\pmb{s}, \pmb{a})$ and $R(\pmb{s}'|\pmb{s}, \pmb{a})$
represents the transition probability and corresponding reward of moving from state $\pmb{s}$ to $\pmb{s}'$ under action $\pmb{a}$. $\pmb{\mu}(\cdot)$ represents the initial state distribution, and $\gamma$ is the discount factor. 

A stationary policy $\pi_{\pmb{\theta}}$ governs the action probability $\pi_{\pmb{\theta}}(\pmb{a}|\pmb{s})$ for a given state $\pmb{s}$, with all policies parameterized by $\pmb{\theta}$. The corresponding infinite-horizon reward is defined as 
\begin{equation}
     J^R({\pi_{\pmb{\theta}}})= \mathbb{E}^\pi_{\pmb{s} \sim \pmb{\mu}}\left[\sum_{t=0}^{\infty} \gamma^t R(\pmb{s}_{t+1} | \pmb{s}_t, \pmb{a}_t)\right].
\end{equation}
The objective of the MDP is to maximize:
\begin{equation}
     \max_{\pi_{\pmb{\theta}} \in \Pi_{\pmb{\theta}}} J^R({\pi_{\pmb{\theta}}}),
\end{equation}
where $\Pi_{\pmb{\theta}}$ is the neural network parameter space.
\subsection{Constrained Markov Decision Processes}
A \ac{cmdp} extends the standard MDP framework by incorporating a set of cost functions, referred to as \textbf{general constraints}. This set is defined as $\mathcal{C} = {(C_i, b_i)}_{i=1}^m$, where each cost function $C_i: \mathcal{S} \times \mathcal{A} \rightarrow \mathbb{R}$ is associated with a corresponding threshold $b_i$. 

The objective of the CMDP is to identify the policy that maximizes the reward while satisfying all specified constraints:
\begin{equation}
\label{eq:cmdp_goal}
       \max_{\pi_{\pmb{\theta}} \in \Pi_{\mathcal{C}}} J^R({\pi_{\pmb{\theta}}}),
\end{equation}
where  $\Pi_{\mathcal{C}} = \Pi_{{\pmb{\theta}}} \cap \left\{\pi_{\pmb{\theta}} \mid J^{C_i}(\pi_{\pmb{\theta}}) \leq b_i, \; \forall i = 1, \ldots, m \right\}$ represents the set of feasible policies that satisfy all constraints and $J^{C_i}(\pi_{\pmb{\theta}})$ is the expected value of a cost function under a policy $\pi$. These constraints can take various forms, such as a discounted cumulative sum, average sum, \etc~\cite{cmdp1999}.

\textbf{Discounted sum constraints} are formulated to ensure that a specified quantity remains below a given threshold over the long term, accounting for cumulative effects rather than instantaneous values at each time step. They are defined as: 
\begin{equation}
\label{eq:discountsum}
 J^{C_i}(\pi_{\pmb{\theta}}) = \mathbb{E}\left[\sum_{t=0}^{T} \gamma^t C_i(\pmb{s}_{t+1} | \pmb{s}_t, \pmb{a}_t)\right] \leq b_i,
\end{equation} 
where $T$ is the episode length. This type of constraint is particularly well-suited for maintaining long-term requirements in robotic operations, such as limiting energy consumption over a certain period. 

%.that does not exceed a predefined threshold.

\textbf{Average sum constraints} ensure that the mean value of a specific variable remains below a given threshold at each time step. Formally, these constraints are expressed as:
\begin{equation} 
\label{eq:avergesum}
 J^{C_i}(\pi_{\pmb{\theta}}) = \mathbb{E}\left[\frac{1}{T}\sum_{t=0}^{T} C_i(\pmb{s}_{t+1} | \pmb{s}_{t}, \pmb{a}_{t})\right] \leq b_i.
 \end{equation}
For instance, these constraints can be utilized to enforce specific motion styles, such as symmetry during walking.

\subsection{Constraints RL Methods}
\subsubsection{\textbf{PPO-Lag}}  
PPO-Lag~\cite{ray2019benchmarking} is one of the most widely used methods in constrained \ac{rl}. The Lagrangian function is:
\begin{equation}
\label{eq:lagloss}
L({\pmb{\theta}}, \lambda_1, \dots,\lambda_i) = J^R(\pi_{\pmb{\theta}}) - \sum_{i=1}^m \lambda_i \left(J^{C_i}(\pi_{\pmb{\theta}}) - b_i\right),
\end{equation}
where \( \lambda_i \) is the Lagrange multipliers. Using this approach, the original constrained optimization problem in \cref{eq:cmdp_goal} is reformulated as an unconstrained optimization problem:
\begin{equation}
\max_{\pi_{\pmb{\theta}} \in \Pi_{\pmb{\theta}}} \min_{\lambda_1, \dots, \lambda_i \geq 0} L({\pmb{\theta}}, \lambda_1, \dots, \lambda_m).
\end{equation}
This optimization process alternates between updating the primal variable \( \pi_{\pmb{\theta}} \) and the dual variables \( \lambda_i \), ultimately yielding an approximate solution to the \cref{eq:cmdp_goal}.

\begin{figure*}[t!]
\centering
\includegraphics[width=\linewidth]{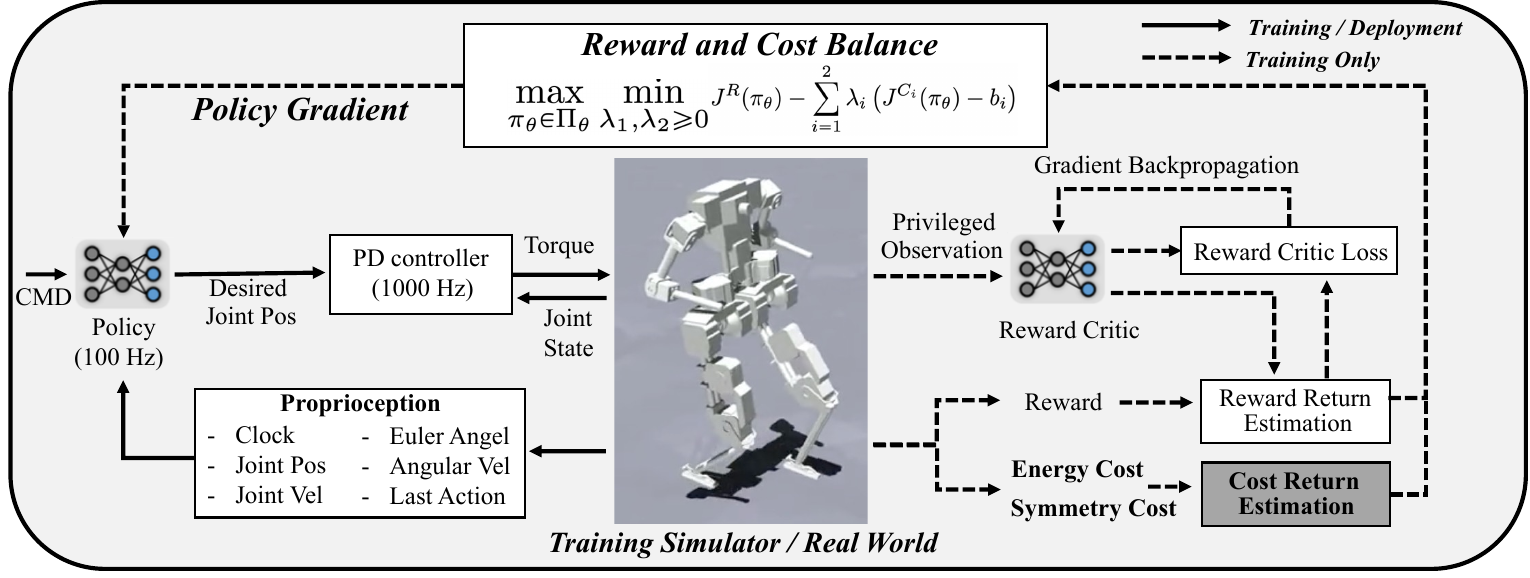}
\caption{\textbf{Overview of the training and deployment process in proposed ECO framework.} The policy network, taking velocity commands and proprioception data as inputs, outputs desired joint positions at $100Hz$ to a PD controller, which updates torque commands at $1kHz$. The reward critic is trained with privileged observations. The simulator provides the reward, energy cost, and symmetry cost, which are used to compute the reward and cost returns. The policy is then updated using the Lagrangian formulation in \cref{eq:lagobject} to balance rewards and costs. The trained policy is directly deployed to the real world.}
\label{fig:framework}
\end{figure*}

\subsubsection{\textbf{CRPO}}
CRPO~\cite{xu2021crpo} alternates between maximizing the reward and minimizing the cost, prioritizing the reduction of costs when constraints are violated. In cases where multiple constraints are violated, a single constraint is randomly selected for optimization:
\begin{equation}
\label{eq:crpoloss}
L_{\text{CRPO}} = \mathds{1}_{n_k=0} \cdot J^R(\pi_{\pmb{\theta}}) - \sum_{i=1}^{n_k}\nu_i \cdot J^{C_i}(\pi_{\pmb{\theta}}),
\end{equation}
where $n_k=\sum_i^{m} \mathds{1}_{J^{C_i}>b_i}$ is the number of violated constraints, $\pmb{\nu}\in \mathds{R}^{n_k}$ is a random one-hot vector used to select a violated constraint to update at each iteration.

\subsubsection{\textbf{IPO}}
IPO~\cite{liu2020ipo} transforms the constrained \ac{rl} problem into an unconstrained one by employing a logarithmic barrier function, defined as follows:
\begin{equation}
\label{eq:ipoloss}
L_{\text{IPO}} = J^R(\pi_{\pmb{\theta}}) - \sum_{i=1}^m \kappa_i^{\text{IPO}} \cdot \text{log}\left(J^{C_i}(\pi_{\pmb{\theta}}) - b_i \right),
\end{equation}
where $\kappa_i^{\text{IPO}} > 0$ is the fixed hyperparameter that controls the steepness of the logarithmic function.

\subsubsection{\textbf{P3O}}  
P3O~\cite{zhang2022penalized} extends the PPO framework by incorporating penalties for constraint violations into the objective function. The objective is defined as:
\begin{equation}
\label{eq:p3oloss}
L_{\text{P3O}} = J^R(\pi_{\pmb{\theta}}) - \sum_{i=1}^m \kappa_i^{\text{P3O}} \cdot \max \{J^{C_i}(\pi_{\pmb{\theta}}) + \epsilon_i ,0\},
\end{equation}
where $\kappa_i^{\text{P3O}}$ represents the fixed weight for each constrain. 
\begin{equation}
\epsilon_i=\mathbb{E}_{(\pmb{s}, \pmb{a}) \sim \text{buffer}}\left[\sum_{t=0}^{T} \gamma^t C_i(\pmb{s}_{t+1} | \pmb{s}_t, \pmb{a}_t)\right] - b_i,
\end{equation}
represents the amount by which the actual cost exceeds the constraint threshold $b_i$, as calculated using data collected from prior interactions with the environment. 

In this paper, we compare these methods and ultimately select PPO-Lag as the constrained optimization approach for ECO. In \cref{sec:framework}, we will provide detailed implementation insights and explain the rationale behind this choice. In \cref{sec:sim}, we will compare their performance to validate this selection.

\section{Proposed ECO Framework}  \label{sec:framework}

In this section, we describe the proposed ECO framework, which aims to minimize energy consumption in humanoid locomotion while ensuring stable and natural leg movements. Both the control policy and reward critic are implemented as fully connected neural networks. As illustrated in \cref{fig:framework}, the control policy takes the robot's proprioception and velocity commands as input and outputs the desired joint positions. We train the policy using \ac{ppo}-Lagrangian~\cite{ji2023safety} in IsaacGym and deploy it directly on the physical robot. The following subsections introduce the state space, action space, constraints settings, reward settings, and the formulation for constraint policy optimization.

\subsection{State Space}
\label{sec:statespace}   %\in \mathbb{R}^{40}
The input to the policy network consists of an observation history comprising $K_f$ consecutive frames, which include the robot's proprioceptive states and velocity commands, with a dimensionality of $\mathcal{S}$. Specifically, $\mathcal{S}$ includes velocity commands $(v^{cmd}_x, v^{cmd}_y, \omega^{cmd}_{yaw}) \in \mathbb{R}^{3}$, clock inputs $(sin(t), cos(t))\in \mathbb{R}^{2}$, joint positions $\pmb{q} \in \mathbb{R}^{n}$, joint velocities $\dot{\pmb{q}} \in \mathbb{R}^{n}$, body angular velocities $\pmb{\omega}_{\text{ang}} \in \mathbb{R}^{3} $, last action $\pmb{a}_{t-1} \in \mathbb{R}^{n}$, and the XY components of body Euler angles $\pmb{\omega}_{\text{ang}} \in \mathbb{R}^{2}$, where $n$ is number of leg \acp{dof}. The body yaw angle is excluded due to its drift on the real hardware.

The state space of our reward critic includes both the inputs used by the policy network and additional privileged information, such as body linear velocity, body yaw angle, external push forces and torques, friction coefficients, body mass, phase identifiers, and contact indicators. These indicators specify the swing and stance phases of each foot, as determined by the contact scheduler. 

\subsection{Action Space}
\label{sec:actionspace}
In this work, we actively control leg joints while the arm joints are fixed to nominal angles. The action from the policy, represented as 
$\pmb{a} \in \mathbb{R}^{n}$, is defined as the deviations from the nominal joint positions $\pmb{q}^\text{nominal}$~\cite{gu2024advancing, lee2024integrating}. This action is fed into the PD controller to convert the action into the desired torque command using the following equation:
\begin{equation} 
\hat{\pmb{\tau}} = \mathbf{K_p}(\pmb{a} + \pmb{q}^\text{nominal} - \pmb{q}) + \mathbf{K_d}(\mathbf{0} - \dot{\pmb{q}}),
\end{equation}
where $\pmb{q},\dot{\pmb{q}}$ denote the current joint position and joint velocity. 

\subsection{Constraints}
\subsubsection{\textbf{Constraints Selection Consideration}}
% In humanoid robot locomotion, three main requirements are typically considered for practical applications: (1) the ability to follow task commands, such as velocity tracking; (2) behavior stability and safety; and (3) energy consumption within specified limits.
Since minimizing energy consumption is the main focus of this paper, 
a naive choice is to mix it with other objectives, such as adherence to task commands, behavior stability, and disturbance rejection. However, regarding energy consumption, we observed that adjusting the reward weights alone for \ac{ppo} does not lead to convergence within the feasible domain (shown in \cref{sec:compare_rl}). Consequently, we select energy consumption as a constraint in this optimization process. Additionally, we integrate reference motion (using Mirror Loss as the reference in this paper) as a constraint to encourage natural motions.

Among prior works~\cite{kim2024not, lee2024exploring}, additional reward terms are often placed within constraints for quadrupedal robots. In \cref{sec:5constraint}, we investigate various constraint settings including the constraints used in~\cite{kim2024not, lee2024exploring}, and find that the current approach struggles with stable convergence as the number of constraints increases. This may be due to humanoid robots having a much smaller constrained feasible domain than quadrupeds, which makes convergence more challenging in the presence of multiple constraints. Therefore, in this work, we focus on two primary constraints: energy minimization and reference motion tracking for humanoid robots.

% For the first two requirements, previous works have demonstrated that reward engineering can effectively address these concerns~\cite{gu2024humanoid, radosavovic2023learning}. However, regarding energy consumption, we observed that adjusting the reward weights alone for \ac{ppo} does not lead to convergence within the feasible domain (shown in \cref{sec:compare_rl}).  Consequently, we select energy consumption as a constraint in this optimization process. Additionally, we integrate reference motion (using Mirror Loss as the reference in this paper) as a constraint, aiming to maintain similarity between the robot's generated and reference motions~\cite{lee2024integrating, gu2024advancing}. Adjusting the constraint threshold allows for explicit control over how closely the robot's behavior follows the reference motion.

% In quadrupedal robots, prior methods often place additional reward terms within the constraints~\cite{kim2024not, lee2024exploring}. In \cref{sec:5constraint}, we will explore experiments with multiple constraints and find that, as the number of constraints increases, the current approach struggles with stable convergence. This may be due to quadrupeds having a larger constrained feasible domain than humanoid robots, which makes convergence more challenging in the presence of multiple constraints. Therefore, in this work, we focus on two primary constraints: energy minimization and reference motion tracking for humanoid robots.

\subsubsection{\textbf{Energy Constraint}}
Due to the challenges in accurately measuring link inertia and quantifying friction and heat dissipation losses, the overall energy consumption of a robot cannot be directly measured. Therefore, in this work, the motor energy consumption, which constitutes the primary part of the total energy consumption, is selected as the evaluation metric. It can be calculated by integrating the absolute power of the $n$ motors.
Since the timestep $dt$ is fixed, the energy cost function is defined as 
\begin{equation}
    C_1(\pmb{s}_{t+1} | \pmb{s}_t, \pmb{a}_t) = \sum_{j=1}^{n} |\pmb{\tau}_t^j \dot{\pmb{q}}_t^j|
\end{equation}  
to exclude $dt$. Here, $\pmb{\tau}_t^j$ and $\dot{\pmb{q}}_t^j$ denote the torque and velocity of the \(j\)-th joint at time \(t\). Using the discounted sum constraint from \cref{eq:discountsum}, the energy constraint is formulated as:
\begin{equation}
\label{eq:jc1}
 J^{C_1}(\pi_{\pmb{\theta}}) = \mathbb{E}\left[\sum_{t=0}^{T} \gamma^t \sum_{j=1}^{n} |\pmb{\tau}_t^j \dot{\pmb{q}}_t^j|\right] \leq b_1,
\end{equation}
where \( b_1 \) is the energy threshold, estimated based on the robot's speed range and walking duration.

\subsubsection{\textbf{Reference Motion Constraint}}
 To enforce stability and mirror symmetry, we define a mirror loss term following~\cite{yu2018learning} as the cost function: 
\begin{equation}
\label{eq:c2}
C_2(\pmb{s}' | \pmb{s}, \pmb{a}) = ||sg\left(\pi_{{\pmb{\theta}}}(\pmb{s}_i)\right) - \Psi_a(\pi_{{\pmb{\theta}}}(\Psi_o(\pmb{s}_i)))||^2,
\end{equation}
where $sg\left(\cdot\right)$ is the stop gradient operator, $\Psi_a(\cdot)$ and $\Psi_o(\cdot)$ are the mirroring functions for actions and states, respectively, and $\pi_{\pmb{\theta}}(\pmb{s}_i)$ is the mean action of the stochastic policy conditioned on the state $\pmb{s}_i$.  This constraint ensures that when the observation input to the policy is mirrored, the corresponding action output is also mirrored. Practically, if the policy learns to lift the left leg at the appropriate moment, this constraint also enforces it to lift the right leg at the corresponding moment, and vice versa. This reference constraint promotes more stable and symmetric behavior within the policy. Such symmetry fosters balanced behavior without the need to prescribe motion trajectories explicitly, thereby reducing the necessity to define trajectory parameters manually.

We apply the average constraint from \cref{eq:avergesum} to this cost function:
\begin{equation}
\label{eq:mirror}
\resizebox{0.89\linewidth}{!}{$ 
 J^{C_2}(\pi_{\pmb{\theta}}) = \mathbb{E}\left[\frac{1}{T}\sum_{t=0}^{T}||\pi_{{\pmb{\theta}}}(\pmb{s}_t) - \Psi_a(\pi_{{\pmb{\theta}}}(\Psi_o(\pmb{s}_t)))||^2 \right] \leq b_2,$}
\end{equation}
where $b_2$ is the constraint threshold of reference motion. 
% Unlike prior approaches~\cite{singh2023learning, yu2018learning}, which incorporate a fixed weight for the mirror constraint in the objective function of \ac{ppo}~\cite{schulman2017proximal}, our method employs constrained optimization to dynamically adjust this coefficient. This ensures that the policy network satisfies the predefined mirror symmetry threshold \(b_2\), achieving greater flexibility and adherence to the constraint.

% (set to 786,432 in our experiments)
\subsection{Rewards}
\label{sec:Rewards}
The aspect of reward design is not the main focus of this paper, therefore we adopt a reward configuration similar to the Humanoid-Gym~\cite{gu2024humanoid}. Since reference motion and energy constraints are directly integrated into the cost function, we have omitted the reference joint position tracking reward, joint velocity penalty reward, and joint torque penalty reward from the original setup. The total reward function is defined as 
\begin{equation}
    R(\pmb{s}_{t+1} | \pmb{s}_t, \pmb{a}_t) = \sum_i \mu_i \cdot r_i,
\end{equation}
where the rewards \(r_i\) and their corresponding scales \(\mu_i\) will be detailed in \cref{sec:sim}.

\begin{table*}[b!]
\centering
\caption{\textbf{Summary of additional reward functions.} The tracking error metric is defined as: 
\(
\phi(e, w) := \exp(-w \cdot \|e\|^2),
\)
where \(e\) represents the tracking error, and \(w\) is the associated weight.  \(\Delta v_{\text{lin}}\) and \(\Delta \omega_{\text{ang}}\) represent the linear and angular velocity errors;  \(\mathds{1}_{(\cdot)}\) is the indicator function for the logical condition (\(\cdot\)), for example, \(\mathds{1}_{\text{stand phase}} = 1 \) when the leg is in stand phase. $t_{air}$ is the duration that the foot remains in the air during locomotion; \(\dot{P}^f\) is the velocity of the feet; \(q_{\text{yaw/roll}}\) is the yaw and roll joint position; \(q_0\) is the default joint position; \(P^b_z\) is the base height; \(\Delta h_f\) is the deviation in foot height; $h_t$ is the target foot height during the swing phase; \(\sum\) represents the summation across all legs. The base height target is set to $0.45m$, the maximum contact force \(F_{\text{max}}\) is set to $50N$, and the target foot height \(h_t\) is $0.03m$. }
\label{tab:additional_rewards}
\resizebox{0.65\linewidth}{!}{%
\begin{tabular}{llr}
\toprule
\textbf{Reward} & \textbf{Equation (\(r_i\))} & \textbf{Scale (\(\mu_i\))} \\ \midrule
Tracking Linear Velocity & \( \phi(\Delta v_{\text{lin}}, 10) \) & 1.2 \\
Tracking Angular Velocity & \( \phi(\Delta \omega_{\text{ang}} , 10) \) & 1.1 \\
Foot Air Time & \( \sum \mathds{1}_{\text{contact}} \cdot \min(t_{\text{air}}, 0.5) \) & 1.0 \\
Foot Contact Velocity & \( \sum \sqrt{\|\dot{P}^f\|} \cdot \mathds{1}_{\text{contact}} \) & -0.05 \\
Foot Clearance & \( \sum \mathds{1}_{|\Delta h_f - h_t| < 0.01} \cdot \mathds{1}_{\text{swing phase}} \) & 1.0 \\
Foot Contact Number & \( \frac{1}{N} \sum (\mathds{1}_{\text{contact}} = \mathds{1}_{\text{stand phase}}) \) & 1.2 \\
Base Orientation & \( \frac{1}{2}(\phi(\text{euler angles}, 10) + \phi(\text{gravity vector}, 20)) \) & 1.0 \\
Foot Contact Forces & \( \sum \max(0, \|F_c\| - F_{\text{max}}) \) & -0.01 \\
Default Joint Pos & \( \phi(q_{\text{yaw/roll}} - q_0, 100) \) & 0.5 \\
Tracking Base Height & \( \phi(P^b_z - 0.45, 100) \) & 0.2 \\
Self-collision Penalty & \( \sum (\mathds{1}_{\text{self-collision}}) \) & -1.0 \\
Action Smoothness & \( \| \pmb{a}_t - \pmb{a}_{t-1}\|^2 + \|\pmb{a}_t-2\pmb{a}_{t-1}+\pmb{a}_{t-2}\|^2 + 0.05 \| \pmb{a}_t\|\) & -0.002 \\
\bottomrule
\end{tabular}}
\label{tab:reward}
\end{table*}

\subsection{Constrained Policy Optimization}

To solve the constrained optimization for energy-efficient humanoid walking, we employ the Lagrangian method to transform the original constrained objective into an unconstrained form, as introduced in \cref{sec:problem}. Having two designed constraints, we formulate the problem as follows:
\begin{equation}
\label{eq:lagloss2}
\max_{\pi_{\pmb{\theta}} \in \Pi_{{\pmb{\theta}}}} \min_{\lambda_1, \lambda_2 \geq 0} J^R(\pi_{\pmb{\theta}}) - \sum_{i=1}^2\lambda_i \bigl(J^{C_i}(\pi_{\pmb{\theta}}) - b_i\bigr).
\end{equation}
Here, $J^{C_1}(\pi_{\pmb{\theta}})$ and $J^{C_2}(\pi_{\pmb{\theta}})$ are cost returns defined by \cref{eq:jc1,eq:mirror}. Solving \cref{eq:lagloss2} involves iteratively updating the policy parameters and the Lagrange multipliers. The update steps alternate between improving the policy based on both reward and cost gradients, and adjusting $\lambda_i$ to enforce constraints. Formally:
\begin{equation}
\label{eq:lagobject}
\resizebox{0.89\linewidth}{!}{$
\begin{aligned}
\pi_{{\pmb{\theta}}^{k+1}} &= \pi_{{\pmb{\theta}}^k} + \alpha(k)\left[\nabla_{{\pmb{\theta}}^k}J^R(\pi_{{\pmb{\theta}}^k}) - \sum_{i=1}^{2}\lambda_i^k \nabla_{{\pmb{\theta}}^k}J^{C_i}(\pi_{{\pmb{\theta}}^k})\right], \\
\lambda_i^{k+1} &= \max\{0, \lambda_i^k - \beta_{i}(k)\bigl(J^{C_i}(\pi_{{\pmb{\theta}}^k}) - b_i\bigr)\}.
\end{aligned}$}
\end{equation}
Here, $\alpha(k)$ and $\beta_i(k)$ are learning rates computed by Adam optimizer~\cite{kingma2014adam} at the training iteration $k$. The projection $\max\{0,x\}$ ensures that each Lagrange multiplier remains non-negative. We integrate this Lagrangian approach into the PPO algorithm~\cite{schulman2017proximal} within an Actor-Critic framework. The policy is optimized to maximize a combined objective:
\begin{equation}
\label{eq:policy}
L_{\text{PPO-Lag}} = L_{\text{R}} - r_t({\pmb{\theta}})\sum_{i=1}^{2}\lambda_i J^{C_i}(\pi_{{\pmb{\theta}}}),
\end{equation}
where $L_{\text{R}}$ is the clipped surrogate objective defined by PPO:

\begin{equation}
\label{eq:jr_ppolag}
\begin{aligned}
L_{\text{R}} &= \min \{r_t\left({\pmb{\theta}}\right) A^{\pi_{\text{old}}}(\pmb{s}_{\leq t}, \pmb{a}_t),\\
&\quad \text{clip}\left(r_t\left({\pmb{\theta}}\right), c_1, c_2\right) A^{\pi_{\text{old}}}(\pmb{s}_{\leq t}, \pmb{a}_t) \},
\end{aligned}
\end{equation}

with the probability ratio 
\begin{equation}
r_t({\pmb{\theta}})=\frac{\pi_{{\pmb{\theta}}}(\pmb{a}_t \mid \pmb{s}_{\leq t})}{\pi_{\text{old}}(\pmb{a}_t \mid \pmb{s}_{\leq t})}.
\end{equation}
In this formulation, $\pi_{\text{old}}$ denotes the behavior policy used for data collection, $A^{\pi_{\text{old}}}$ is the advantage computed via GAE~\cite{schulman2015high}, and $c_1,c_2$ define the PPO clipping range.

Similarly, within the PPO framework, the final objectives of IPO, P3O, and CRPO are as follows:
\begin{equation}
\label{eq:baseline_loss}
\resizebox{0.89\linewidth}{!}{$
\begin{aligned}
L_{\text{IPO}} &=  L_{\text{R}} - r_t\left({\pmb{\theta}}\right) \cdot \sum_{i=1}^2 \kappa_i^{\text{IPO}} \cdot \text{log}\left(J^{C_i}(\pi_{\pmb{\theta}}) - b_i \right),\\
L_{\text{P3O}} &= L_{\text{R}} - r_t\left({\pmb{\theta}}\right) \cdot \sum_{i=1}^2 \kappa_i^{\text{P3O}} \cdot \max \{J^{C_i}(\pi_{\pmb{\theta}}) + \epsilon_i ,0\},\\
L_{\text{CRPO}} &= \mathds{1}_{n_k=0} \cdot  L_{\text{R}} - r_t\left({\pmb{\theta}}\right) \cdot \sum_i^2 \nu_i \cdot J^{C_i}(\pi_{\pmb{\theta}}).
\end{aligned}$}
\end{equation}

Among the algorithms compared, P3O (with fixed \(\kappa_i^{\text{P3O}} = 1.0\)) and PPO-Lag (\textbf{with initial} \(\lambda_i = 0\), \(\beta_i = 1e{-}3\)) required minimal parameter tuning and achieved low constraint violations. While P3O demonstrated comparable energy performance, PPO-Lag exhibited better convergence speed and stability, thereby selected for implementing the constrained policy optimization approach in ECO.  %Experimental results are provided in \cref{sec:sim}.

The reward critic is trained with the following loss~\cite{rudin2022learning}:
\begin{equation}
\label{eq:critic}
 \resizebox{0.89\linewidth}{!}{$
\begin{aligned}
& L_{\text{Reward Critic}} = \max \{\|V(\pmb{s}_t) - R_t\|_2, \\
&\|V_{\text{target}}(\pmb{s}_t) + \text{clip}(V_{\text{target}}(\pmb{s}_t) - V(\pmb{s}_t), c_1, c_2) - R_t\|_2 \},
\end{aligned}$}
\end{equation}
where $R_t$ is the cumulative return at time $t$, $V(\pmb{s}_t)$ denotes the value estimated by the critic at state $\pmb{s}_t$, and $V_{\text{target}}(\pmb{s}_t)$ is the value approximated by the target critic at state $\pmb{s}_t$. Unlike previous studies~\cite{ji2023safety, ray2019benchmarking} that utilize a cost critic, we estimate cost return for energy using the Monte Carlo approach~\cite{williams1992simple}, which achieves comparable performance while requiring fewer network parameters and minimal design choices.

\section{Simulation} \label{sec:sim}

\subsection{Robot Platform}
We use the kid-sized humanoid robot BRUCE~\cite{liu2022design, zhang2023design, wr-bruce} as an example to demonstrate the performance of the proposed ECO framework. BRUCE is $70cm$ tall, weighs $4.8kg$, and has a total of 16 \acp{dof}, with 3 per arm and 5 per leg. All leg \acp{dof} are driven by liquid-cooled proprioceptive actuators with peak torque $10.5Nm$, while the arm \acp{dof} are driven by Dynamixel servo motors. The PD controller gains for both legs are set as $\mathbf{K_p}=\textit{diag}(7, 10, 7, 10, 1.5)$ and $\mathbf{K_d}=\textit{diag}(0.2, 0.4, 0.2, 0.4, 0.08)$ for the hip yaw, hip pitch, hip roll, knee pitch, and ankle pitch joints, respectively.

\subsection{Learning Setups}

\subsubsection{\textbf{Traning Setups}} The policy training is conducted in the IsaacGym~\cite{isaacgym} with parallelization across 8192 environments. The utilized rewards and scales are summarized in \cref{tab:additional_rewards}. The constraint policy optimization of ECO utilizes the PPO-Lagrangian~\cite{ray2019benchmarking} and the hyperparameters are shown in \cref{tab:hyperparameters}. We employ domain randomization~\cite{pinto2017asymmetric} to enhance transfer to unseen scenarios, with specific parameters listed in \cref{tab:domain_randomization}. %We found that noise in the initial Euler angles and joint velocities significantly improves sim-to-real transfer stability. 
During training, two types of external perturbations are applied to enhance the policy robustness: force disturbances every $2s$, lasting $0.001s$ with magnitudes randomly sampled between $[-100, 100]N$, and velocity impulses every $4s$, also lasting $0.001s$, with linear velocities up to $0.2m/s$ and angular velocities up to $0.4rad/s$. The simulation is executed at a frequency of $1kHz$, with a control frequency of $100Hz$. The training process is carried out on a system equipped with an Intel Core i9-13900KFN processor and NVIDIA GeForce RTX 4090, and each training run consists of 2000 policy iterations. Convergence is typically achieved within approximately $4h$ of wall-clock time. We use the policy trained after $10h$ for sim-to-real experiments.

\begin{table}[t!]
\centering
\caption{\textbf{Overview of domain randomization.} The domain randomization terms and the associated parameter ranges are listed. Additive randomization increments the parameter by a value within the specified range while scaling randomization adjusts it by a multiplicative factor from the same range.}
\label{tab:domain_randomization}
\resizebox{\linewidth}{!}{%
\begin{tabular}{lcccc}
% \hline
\toprule
\textbf{Parameter} & \textbf{Unit} & \textbf{Range} & \textbf{Operator} & \textbf{Type} \\
\hline
Payload & kg & [-0.5, 0.5] & additive & Uniform \\
Body COM Displacement  & m & [-0.05, 0.05] & additive & Uniform \\
Floor Friction & - & [0.1, 2.0] & - & Uniform \\
Restitution & - & [0.0, 0.5] & additive & Uniform \\
Motor Strength & \% & [0.9, 1.1] & scaling & Uniform \\
Joint Friction & \% & [0.02, 0.05] & scaling & Uniform \\
Joint armature & \% & [0.0, 0.01] & additive & Uniform \\
Action Delay & ms & [0, 10] & - & Uniform \\
Kp Factors & \% & [0.9, 1.1] & scaling & Uniform\\
Kd Factor & \% & [0.9, 1.1] & scaling & Uniform\\
Initial Euler Angle Noise& rad & [-0.1, 0.1] & additive & Uniform  \\
Joint Position Noise & rad & [-0.03, 0.03] & additive & Gaussian ($1 \sigma$) \\
Joint Velocity Noise & rad/s & [-0.6, 0.6] & additive & Gaussian ($1 \sigma$) \\
% Base Lin. Vel. & m/s & [0.0, 0.15] & additive \\
Angular Velocity Noise& rad/s & [-0.06, 0.06] & additive & Gaussian ($1 \sigma$) \\
Euler Angle Noise & rad & [-0.018, 0.018] & additive &  Gaussian ($1 \sigma$)  \\
% Action Delay & - & [0, 1] & - \\
% Motor Offset & rad & [-0.05, 0.05] & additive \\
% Motor Damping & Nms/rad & [0.3, 4.0] & scaling \\
% Gravity & m/s\textsuperscript{2} & [0.0, 0.67] & additive \\
\bottomrule
\end{tabular}}
\end{table}

\begin{table}[t!]
\centering
\caption{\textbf{Hyperparameters in the training process.}}
\label{tab:hyperparameters}
\resizebox{0.84\linewidth}{!}{%
\begin{tabular}{lc}
\toprule
\textbf{Parameter} & \textbf{Value} \\
\hline
% Number of GPUs & 1 RTX A6000 \\
Number of Environments & 8192 \\
Number Training Epochs & 2 \\
% Steps per Environment & 49152\\
Batch size & $8192 \times 96$ \\
Episode Length & $24s$ \\
Discount Factor of Reward& 0.994 \\
Discount Factor of Energy Cost & 0.90 \\
PPO Clip Coefficient $c1,c2$ & 0.8, 1.2 \\
\hline
Frame Stack of Single Observation $K_f$ & 15 \\
Frame Stack of Single Privileged Observation & 3 \\
Number of Single Observation $\mathcal{S}$ & 40 \\
Number of Single Privileged Observation & 55 \\
\hline
Initial Lagrangian Multipliers $\lambda_1,\lambda_2$  & 0.0, 0.0\\
Learning Rate of Lagrangian Multiplier$\beta_1,\beta_2$  & 0.001,0.001\\
Cost Threshold $b_1$ ($0.1m/s$) & $60J = 2.5W \times24s$ \\
Cost Threshold $b_1$ ($0.15m/s$) & $70J \approx 2.9W\times24s$ \\
Cost Threshold $b_1$ ($0.2m/s$) & $80J\approx3.3W\times24s$ \\
Cost Threshold $b_2$ & $0.05$\\
% \hlne
\bottomrule
\end{tabular}}
\end{table}

\subsubsection{\textbf{Constraint Thresholds Search}}
Since the PPO baseline policy described in \cref{sec:baselines} converges to a suboptimal solution, it serves as a useful initial reference for selecting constraint thresholds. For the reference motion constraint, we empirically observed that the mirror-symmetry loss value obtained from the PPO baseline ensures robust and symmetric walking behavior. Accordingly, we set the threshold $b_2$ in \cref{eq:mirror} as 0.05 across all experiments. For the energy constraint threshold, we adopt a linear search strategy~\cite{kim2024not} to determine appropriate values, further details are provided in \cref{sec:energy_constraint_threshold}. Each threshold
search takes around 4 hours of wall-clock time depending on the chosen step size. Additionally, as summarized in \cref{tab:hyperparameters}, energy consumption thresholds may vary with different target velocities.

\subsection{Baselines}
\label{sec:baselines}
For comparison, five baseline approaches are chosen to demonstrate that our proposed ECO framework significantly improves energy efficiency without degenerating the performance of humanoid robots in walking tasks. 
The \textbf{MPC baseline} uses the state-of-the-art model-based hierarchical method which combines a high-level MPC planner with a simplified model and a low-level whole-body controller with the full-body dynamics~\cite{shen2022implementation}. Humanoid-Gym~\cite{gu2024humanoid} is used as the \textbf{model-free RL baseline (PPO)} for training end-to-end locomotion policies in humanoid robots. To facilitate a rigorous comparison with the ECO model, the PPO baseline has been augmented with an energy penalty reward~\cite{fu2021minimizing}, which serves as the constraint within the ECO framework. {We selected scales of energy penalty reward \(\mu_e \in \{-0.001, -0.01, -0.1, -0.03, -0.05, -0.08, -0.015, -0.02, \\-0.025\}\), while all other rewards and scales are aligned with those of ECO as detailed in \cref{tab:additional_rewards}.} Additionally, a mirror loss is included in the PPO loss function to encourage symmetry, with a fixed weight of 0.1. Three constrained RL baseline methods are chosen as introduce in \cref{eq:baseline_loss}. For IPO and P3O, we fine-tune the hyperparameters as follows: \(\kappa_1^{\text{P3O}} = 1.0\), \(\kappa_2^{\text{P3O}} = 1.0\), \(\kappa_1^{\text{IPO}} \in \{1.0, 10.0\}\), and \(\kappa_2^{\text{IPO}} = 1.0\).

\begin{figure*}[t!]
\centering
\includegraphics[width=0.7\linewidth,trim=0cm 0.2cm 0cm 0cm,clip]{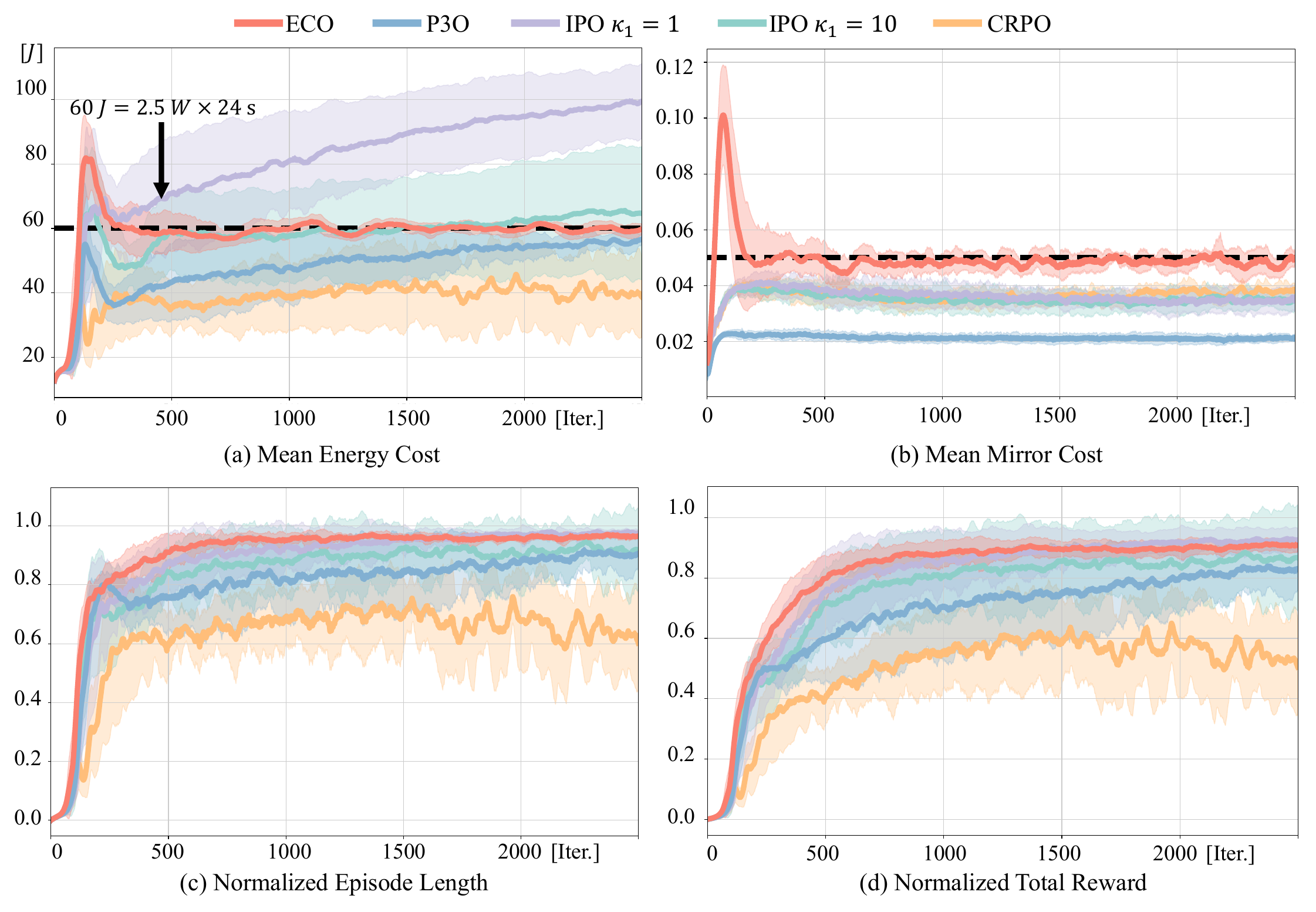}
\caption{\textbf{Comparison of training metrics for ECO, P3O, IPO, and CRPO.} The energy consumption and mirror reference motion thresholds are set at $60J$ and 0.05, respectively, as indicated by the \textbf{black dashed lines} in (a) and (b). Results averaged over 10 random seeds.}
\label{fig:main_exp}
\end{figure*}

\begin{figure*}[t!]
\centering
\includegraphics[width=0.7\linewidth,trim=0cm 0.3cm 0cm 0.4cm,clip]{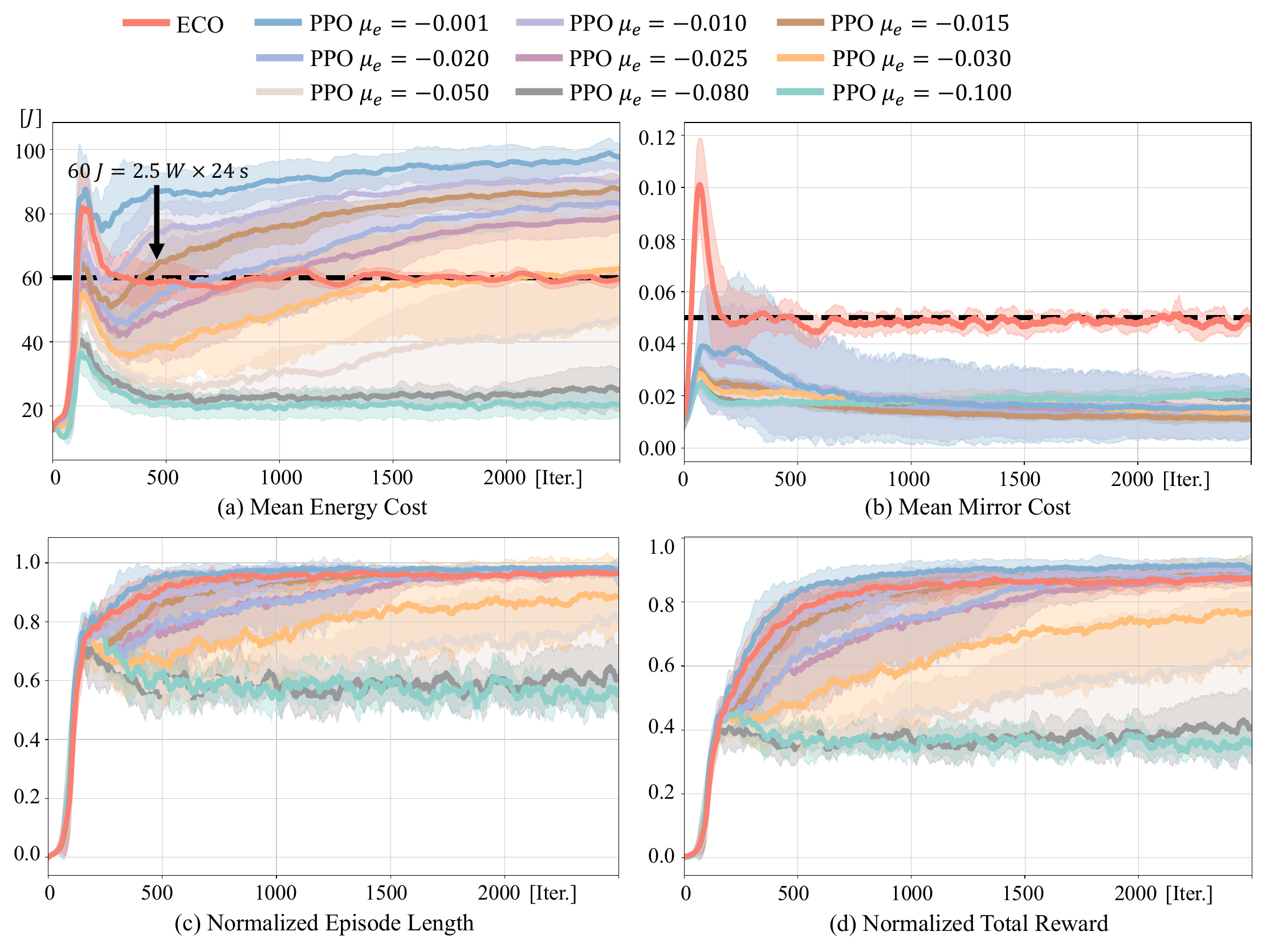}
\caption{\textbf{Comparison of training metrics for ECO and the PPO.} The energy consumption and mirror reference motion thresholds are set at $60J$ and 0.05, respectively, as indicated by the \textbf{black dashed lines} in (a) and (b). Results averaged over 10 random seeds.}
\label{fig:main_ppo}
\end{figure*}

\begin{figure*}[t!]
\centering
\includegraphics[width=0.9\linewidth,trim=0.4cm 11.2cm 1.4cm 0.5cm,clip]{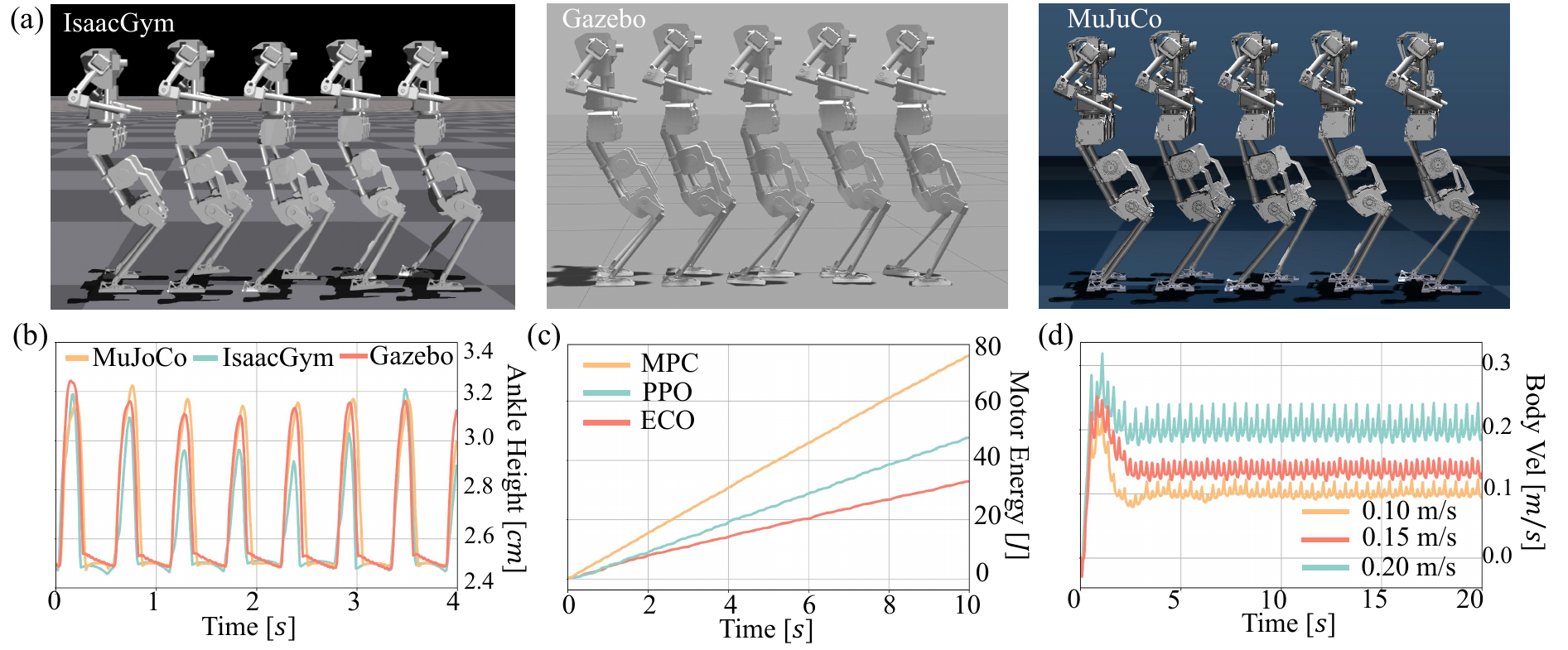}
\caption{\textbf{Sim-to-Sim transfer results.} For visual clarity, single-run deployment curves are shown. (a) Visual comparison of walking stability across simulators; (b) Ankle height consistency across simulators; (c) Motor energy consumption at $0.1m/s$ over $10s$ in Gazebo; (d) Body velocity tracking in MuJoCo in different speed commands.}
\label{fig:sim2sim}
\end{figure*}

\subsection{Simulation Results}
\subsubsection{\textbf{Comparing Constraint RL Algorithms}}
\label{sec:compare_rl}
We compare the proposed ECO framework against PPO, IPO, P3O, and CRPO under a $0.1m/s$ linear velocity tracking task, subject to energy and reference motion constraints. To assess the performance, 
we evaluate four key metrics, each capturing a distinct aspect of the task:
(1) Normalized total reward $\hat{J}^R$ — reflects overall task performance;
(2) Normalized episode length $T_{\text{ep}}$ — measures stability and robustness;
(3) Mean energy cost $\hat{J}^{C_1}$ — quantifies energy efficiency;
(4) Mean mirror cost $\hat{J}^{C_2}$ — evaluates motion symmetry.
The metrics are formally defined as:
\begin{align}
\hat{J}^{R}&=\frac{\frac{1}{E} \sum_{i = 1}^{E} \sum_{t = 0}^{T} r^{t,i} - R_{\min}}{R_{\max} - R_{\min}},\\
T_{\text{ep}}&=\frac{\frac{1}{E} \sum_{i = 1}^{E} T_{\text{ep}}^i - T_{\text{ep}}^{\max}}{T_{\text{ep}}^{\max} - T_{\text{ep}}^{\min}},\\
\hat{J}^{C_1}&=\frac{1}{E} \sum_{i = 1}^{E} \sum_{t = 0}^{T} C_1(\pmb{s}_t^i,\pmb{a}_t^i),\\
\hat{J}^{C_2}&= \frac{1}{B}\sum_{i=1}^{B} C_2(\pmb{s}_i,\pmb{a}_i),
\end{align}
where $r^{t,i}$ is the reward at time $t$ of episode $i$, and $R_{\min}$, $R_{\max}$ are the minimum and maximum cumulative rewards observed across all algorithms, $T_{\mathrm{ep}}^{i}$ is the length of episode $i$, and $T_{ep}^{min}$, $T_{ep}^{max}$ are the minimum and maximum episode lengths among all methods, $(\pmb{s}_i,\pmb{a}_i)$ is the $i$-th state–action sample in a batch and $B$ is the batch size. We enforce the average‐sum constraint over each gradient batch as in~\cite{singh2023learning}, using the same batch size across all algorithms.

% \begin{itemize}[leftmargin=1em]
% 	\item Normalized total reward over $E$ episodes: $\frac{\frac{1}{E} \sum_{i = 1}^{E} \sum_{t = 0}^{T_{\text{ep}}} r^{t,i} - R_{min} }{R_{max} - R_{min}}$,  
%   
% 	\item Normalized episode length over $E$ episodes: $\frac{\frac{1}{E} \sum_{i = 1}^{E} T_\text{ep}^i - T_{ep}^{max}}{T_{ep}^{max}-T_{ep}^{min}}$, where 
% 	\item Mean energy cost over $E$ episodes: $\frac{1}{E} \sum_{i = 1}^{E} \sum_{t = 0}^{T_\text{ep}} C_1(s_t^i,a_t^i)$.
% 	\item Mean mirror cost over a gradient-update batch of size $B$: $\frac{1}{B}\sum_{i=0}^{B} C_2(s_i,a_i)$, where each 
% \end{itemize}
% }
As shown in \cref{fig:main_exp}, ECO (initialized with $\lambda_i = 0$, $\beta_i = 10^{-3}$) consistently converges to the target energy threshold of $60J$ while maintaining a normalized episode length close to 1.0, demonstrating both energy efficiency and training stability. In comparison, P3O satisfies both the energy and reference motion constraints but requires more iterations to converge. IPO exhibits sensitivity to the penalty coefficient $\kappa_1^{\text{IPO}}$: with $\kappa_1^{\text{IPO}} = 1.0$, it fails to satisfy the energy constraint; increasing it to $\kappa_1^{\text{IPO}} = 10.0$ enforces the constraint but leads to reduced episode length due to overly conservative behavior. CRPO, while capable of satisfying the constraints, exhibits unstable convergence, likely due to frequent policy shifts between optimizing for reward and constraint cost.

As shown in \cref{fig:main_ppo}, we explore three orders of magnitude for the energy reward coefficient in PPO ($\mu_e$), but none yields a satisfactory balance between energy efficiency and stability. Lower coefficients failed to reduce energy consumption, while higher ones caused instability and frequent falls. These results highlight the difficulty of manually tuning reward weights in multi-objective PPO. In contrast, ECO directly enforces energy and motion constraints, avoiding such tuning and achieving robust, energy-efficient behavior.

Regarding the reference motion constraint (\cref{fig:main_exp}(b)), IPO, P3O, and CRPO rapidly converge to the predefined threshold of $0.05$, though their resulting behaviors are more conservative than those of ECO. In contrast, PPO with various energy coefficients converges to different levels of mirror cost (\cref{fig:main_ppo}(b)), often yielding asymmetric and unstable gaits. These results highlight that constrained RL offers a more principled and efficient framework for aligning policy behaviors with target objectives without heavy hyperparameter tuning.

Finally, we note that PPO, P3O, and IPO rely on fixed constraint coefficients ($\mu_e$, $\kappa^{\text{P3O}}$, $\kappa^{\text{IPO}}$), and CRPO lacks constraint-specific parameters altogether. These approaches are therefore prone to convergence issues or require careful manual tuning. In contrast, ECO leverages PPO-Lag to dynamically update the Lagrange multipliers $\lambda_i$ using the Adam optimizer, allowing adaptive scaling the impact of constrains during training. This dynamic adjustment significantly improves convergence stability and highlights PPO-Lag as a more effective strategy for optimizing energy-efficient humanoid locomotion.

%Additionally, our approach achieves greater energy efficiency by significantly reducing foot air time, as illustrated in \cref{fig:main_exp}(c). The contact forces upon landing are also consistently lower than those observed with the PPO-based methods, as depicted in \cref{fig:main_exp}(d).

\subsubsection{\textbf{Sim-to-Sim Transfer}}
To evaluate the robustness and generalizability of the learned policy, we transferred it to two additional simulation environments, MuJoCo~\cite{todorov2012mujoco} and Gazebo~\cite{gazebo}, to assess ECO's sim-to-sim performance and benchmark it against baseline methods.
As shown in \cref{fig:sim2sim}(a), our method achieves stable walking across three simulation environments. \cref{fig:sim2sim}(b) and (c) demonstrate that ECO consistently maintains a similar gait pattern during cross-simulator transfer and reduces energy usage in Gazebo while BRUCE walks at $0.1m/s$ over a $10s$ period. \cref{fig:sim2sim}(d) further shows that our method accurately tracks linear velocity commands at three different target speeds ($0.1m/s, 0.15m/s, 0.2m/s$) in MuJoCo. We report summary statistics over 10 runs in \cref{tab:cross_simulator}. Results show that MPC incurs the highest energy cost and fails to track higher target speeds (e.g., $0.2m/s$), likely due to its QP solver being unable to resolve contact dynamics in finite time. PPO exhibits better performance than MPC but with moderate energy consumption. ECO, in contrast, not only achieves accurate velocity tracking but also maintains significantly lower energy consumption—approximately 3 times lower than MPC and 1.4 times lower than PPO.

\begin{figure}[b!]
\centering
\includegraphics[width=\linewidth,trim=0.1cm 0cm 0cm 0.1cm,clip]{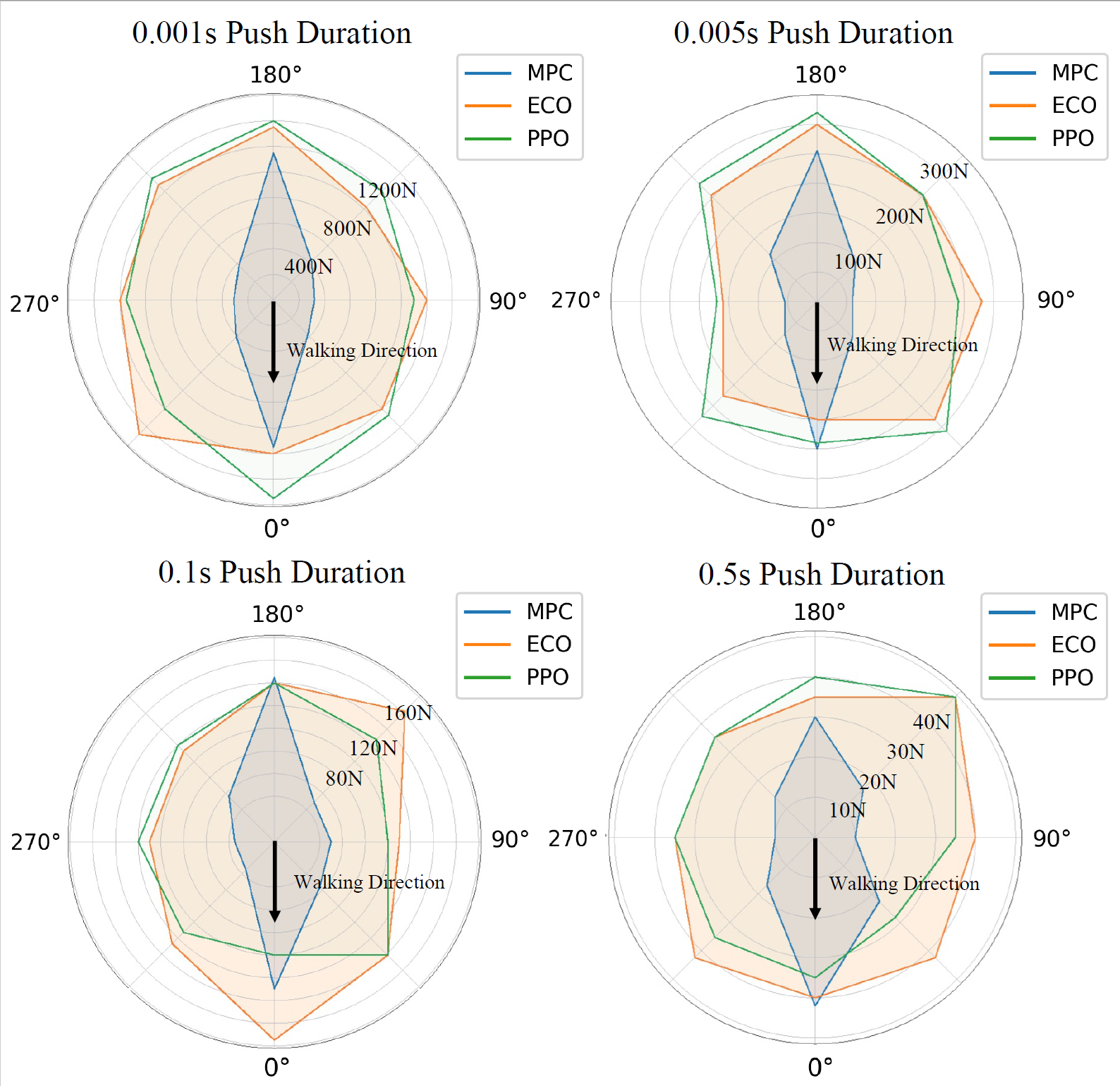}
\caption{\textbf{Benchmark of disturbance rejection.} 
BRUCE was subjected to pushes of varying durations from 8 different directions (\(0^{\circ}, 45^{\circ}, 90^{\circ}, 135^{\circ}, 180^{\circ}, 225^{\circ}, 270^{\circ}, 315^{\circ}\)). Results are averaged over 10 runs.
%The maximum force at which BRUCE failed to maintain balance while walking is recorded for comparison between ECO and two baselines.
}
\label{fig:dist_rej}
\end{figure}

\begin{table*}[ht!]
\caption{\textbf{Velocity tracking and energy consumption across simulators.} Aggregated over 10 runs, each lasting 10 seconds.}
\centering
\resizebox{0.85\textwidth}{!}{ % adjust width here
\begin{tabular}{lcccccc}
\toprule
\textbf{Method} & \multicolumn{2}{c}{\textbf{Target:} ($0.1m/s$, $2.5W$)} & \multicolumn{2}{c}{\textbf{Target:} ($0.15 m/s$, $2.9W$)} & \multicolumn{2}{c}{\textbf{Target:} ($0.2m/s$, $3.3W$)} \\
\cmidrule(lr){2-3} \cmidrule(lr){4-5} \cmidrule(lr){6-7}
& Speed (m/s) & Energy (W) & Speed (m/s) & Energy (W)  & Speed (m/s) & Energy (W) \\
\midrule
PPO (Isaacgym)  & $ 0.1129 \pm \mathbf{0.00071}$ & $3.1301\pm\mathbf{0.0165}$ & $\mathbf{0.1534} \pm 0.0030$ & $4.0965\pm0.0591$& $0.2141 \pm 0.0043$ & $4.6311\pm0.0666$ \\
\rowcolor{Gray!15} \textbf{Ours (Isaacgym)}  & $\mathbf{0.1026} \pm 0.00279$ & $\mathbf{2.1952}\pm0.0458 $ & $0.1617 \pm \mathbf{0.00188}$ & $\mathbf{2.2401}\pm\mathbf{0.0239}$ & $\mathbf{0.2131} \pm \mathbf{0.0041}$ & $\mathbf{2.8474}\pm\mathbf{0.0641}$ \\

\midrule
PPO (Mujoco)   & $0.1092 \pm \mathbf{0.0005}$ & $2.8150\pm 0.00895$ & $\mathbf{0.1567} \pm 0.00042$ & $3.7186\pm0.00957$ & $ \mathbf{0.2028} \pm 0.00089$ & $4.0692\pm0.0162$ \\
\rowcolor{Gray!15} \textbf{Ours (Mujoco)}   & $\mathbf{0.10033} \pm 0.00064$ & $\mathbf{2.00009} \pm \mathbf{0.0066}$ & $0.1572 \pm  \mathbf{0.00022}$ & $\mathbf{2.7098}\pm\mathbf{0.00561}$ & $0.2038 \pm \mathbf{0.00064}$ & $\mathbf{2.8497}\pm\mathbf{0.00838}$ \\

\midrule
PPO (Gazebo)   & $\mathbf{0.1062} \pm  \mathbf{0.0011}$ &  $3.1508 \pm \mathbf{0.0154}$ & $\mathbf{0.1528} \pm \mathbf{0.00092}$ & $4.3922\pm 0.0181$ & $\mathbf{0.2027}\pm 0.0037$ & $ 5.1826 \pm \mathbf{0.0434}$ \\
MPC (Gazebo)\textsuperscript{*}  & $0.1075 \pm 0.0028$ & $9.2693\pm 0.1604 $ & $0.1322 \pm 0.0047$ & $9.9155 \pm 0.3843$ & $0.1523 \pm 0.0057$ & $10.8626 \pm 0.3497$ \\
\rowcolor{Gray!15} \textbf{Ours (Gazebo)}   & $0.0930 \pm 0.0029$ &  $\mathbf{2.3653} \pm 0.0201$ & $0.1542 \pm  0.0013$ & $\mathbf{2.9737}\pm \mathbf{0.0133}$ & $0.2159\pm \mathbf{0.0035}$ & $ \mathbf{3.5067} \pm 0.0559$ \\
\bottomrule
\label{tab:cross_simulator}
\end{tabular}}
\\
\vspace{-1.5ex}
\scriptsize{\textsuperscript{*} MPC baseline method is exclusively implemented in Gazebo so that its performances in Isaacgym and Mujoco are disregarded.}

\label{tab:walk_trot_bounce}
\end{table*}

\subsubsection{\textbf{Disturbance Rejection Test}}
To verify that the walking robustness has not been sacrificed in the learned policy, we conduct disturbance rejection tests on BRUCE while walking in Gazebo. External forces are applied at the origin of the coordinate frame attached to BRUCE’s body link. The timing is chosen as vulnerable transitions from a double-stance phase to lifting the left foot or the right foot. Since the response to perturbations varies with the direction in these two cases, we record the maximum force BRUCE could resist for both timings and use the smaller of the two values for analysis. As shown in \cref{fig:dist_rej}, the proposed ECO framework demonstrates greater robustness across all directions, especially in the sagittal direction, compared to the baseline MPC, and shows comparable robustness to PPO. Although ECO has not encountered pushes with a duration longer than $0.001s$ in the training stage, ECO still performs well under longer pushes, indicating that our method has not overfitted to improve energy efficiency in compromise of robustness. Besides controlled pushes, two qualitative disturbance experiments in Mujoco are conducted to evaluate robustness under more diverse and unpredictable scenarios: (1) different links of BRUCE are manually dragged at random timings with random magnitudes, producing varied disturbance directions and intensities; (2) Balls are thrown to BRUCE with random velocities to create different impulsive contacts. As shown in the accompanying video, ECO successfully maintains stable walking and recovers from these random perturbations, further demonstrating that the learned policy generalizes beyond the specific disturbance patterns used in training.

\begin{figure}[t!]
\centering
\includegraphics[width=0.95\linewidth]{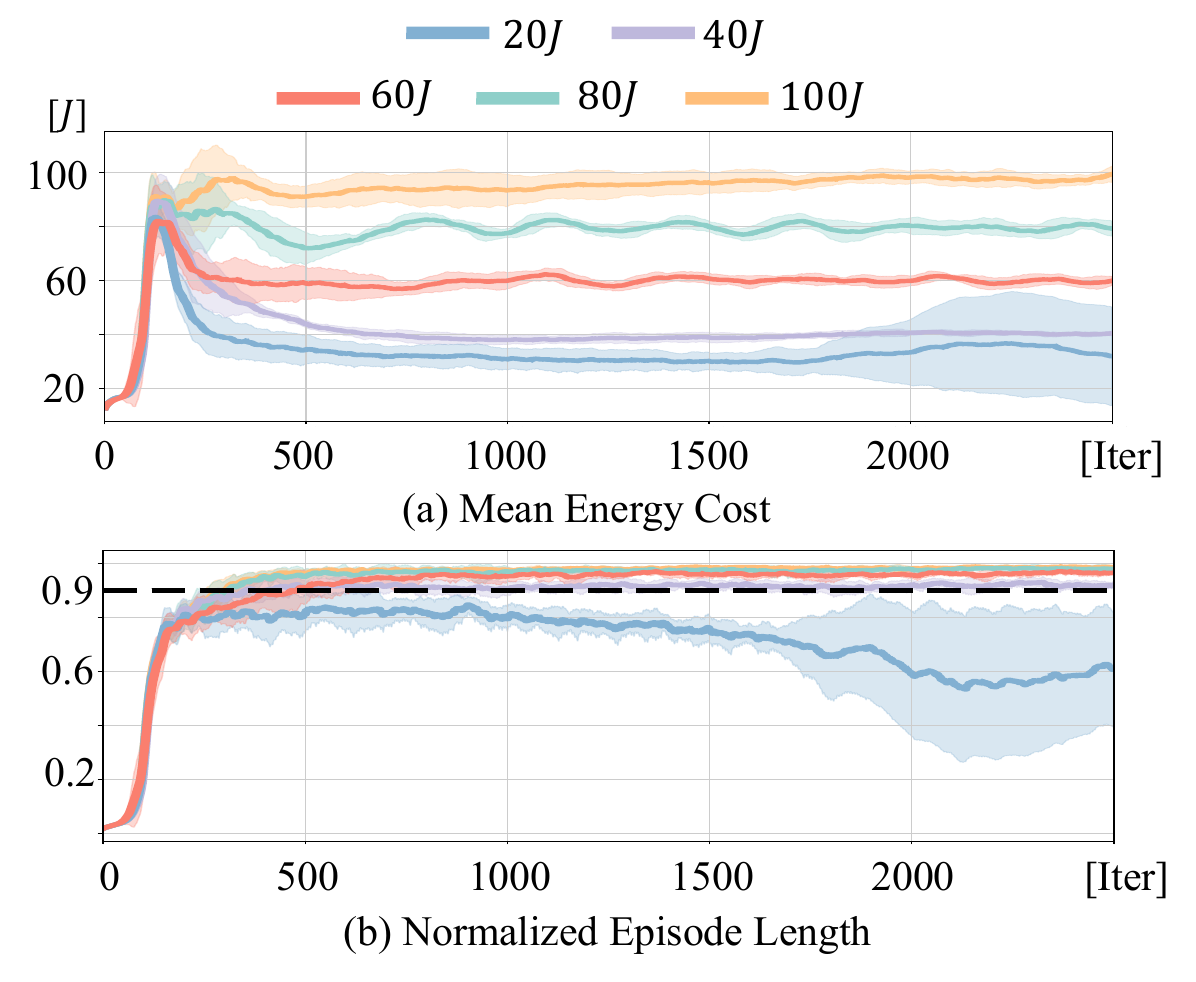}
\caption{\textbf{Energy constraint threshold analysis.} (a)(b): The mean energy cost and normalized episode length for different energy thresholds. Results averaged over 10 random seeds.}
\label{fig:linear_search}
\end{figure} 

\subsubsection{\textbf{Energy Constraint Threshold Analysis}}
\label{sec:energy_constraint_threshold}
We observe that the baseline method PPO converges (achieving a normalized episode length of $0.9$) within $1500$ training iterations while consuming approximately $100J$ of energy with the target walking velocity as $0.1m/s$. Based on this, we conduct a linear search over energy thresholds of $20J,40J,60J,80J$, and $100J$ to identify the minimum threshold that enables the policy to achieve similar normalized episode length within 1500 iterations, as shown in \cref{fig:linear_search}. The $60J$ threshold provides the best trade-off between energy efficiency and locomotion stability. This approach is effective as it leverages physically meaningful thresholds while supporting parallel evaluation. 
%As illustrated in \cref{fig:linear_search}(c), 
A finer-grained search (e.g., with $10J$ increments) can discover policies with even lower energy consumption (e.g., $50J$) that still maintain stability. If further energy reduction is required, smaller step sizes enable finer control at the cost of additional computation.

\begin{figure}[t!]
\centering
\includegraphics[width=0.85\linewidth,trim=0.6cm 0.3cm 0cm 0cm,clip]{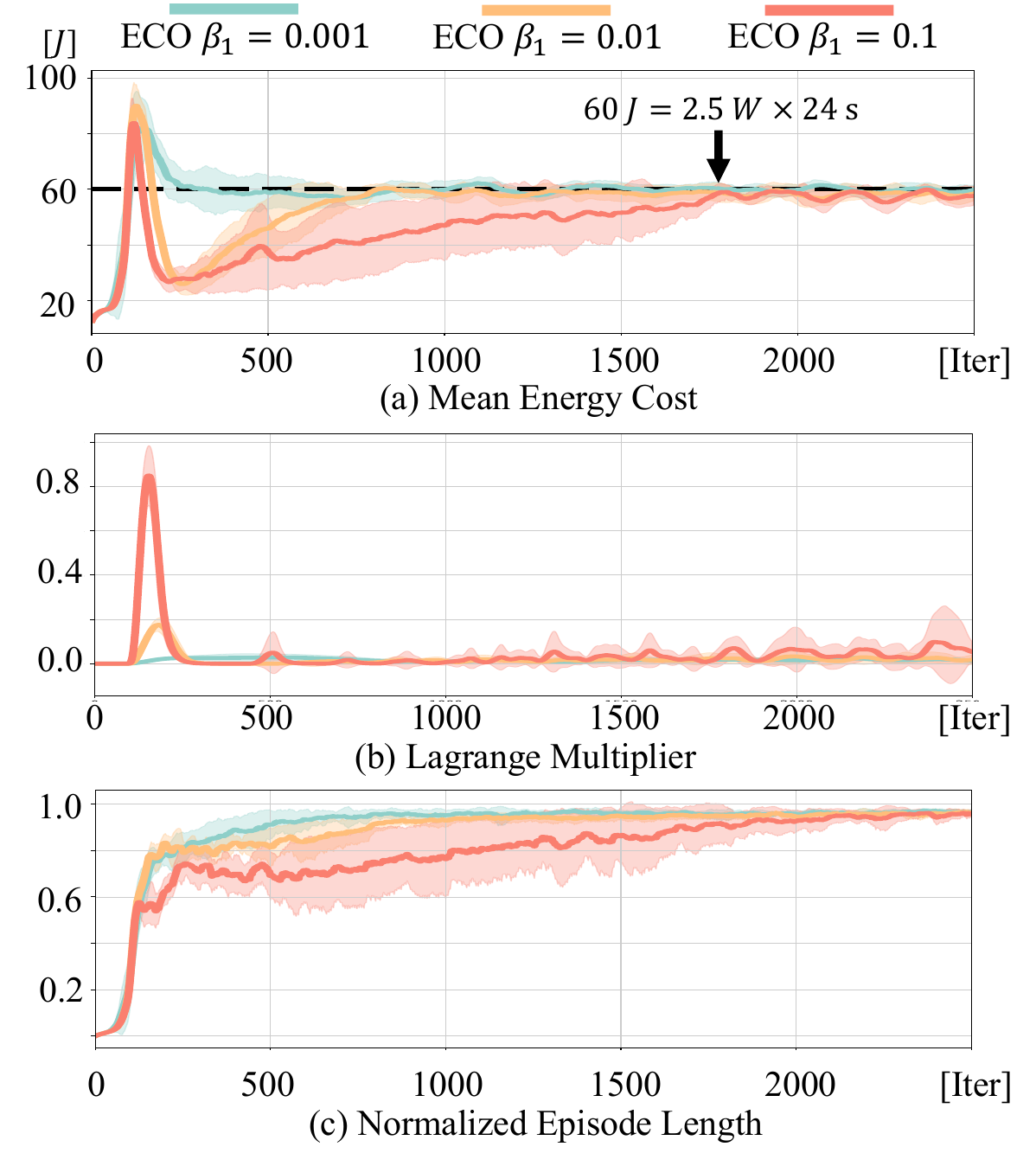}
\caption{\textbf{Hyperparameter sensitivity analysis.} The mean energy cost (a), Lagrange multiplier $\lambda_1$ (b), and normalized episode length (c) of ECO with the energy constraint $b_1=60J$, evaluated across different learning rates for $\lambda_1$ (\(\beta_1 = 0.001, 0.01, 0.1\)) during training. Results averaged over 10 random seeds.}
\label{fig:hyper}
\end{figure} 

\begin{figure*}[t!]
\centering
\includegraphics[width=0.95\linewidth]{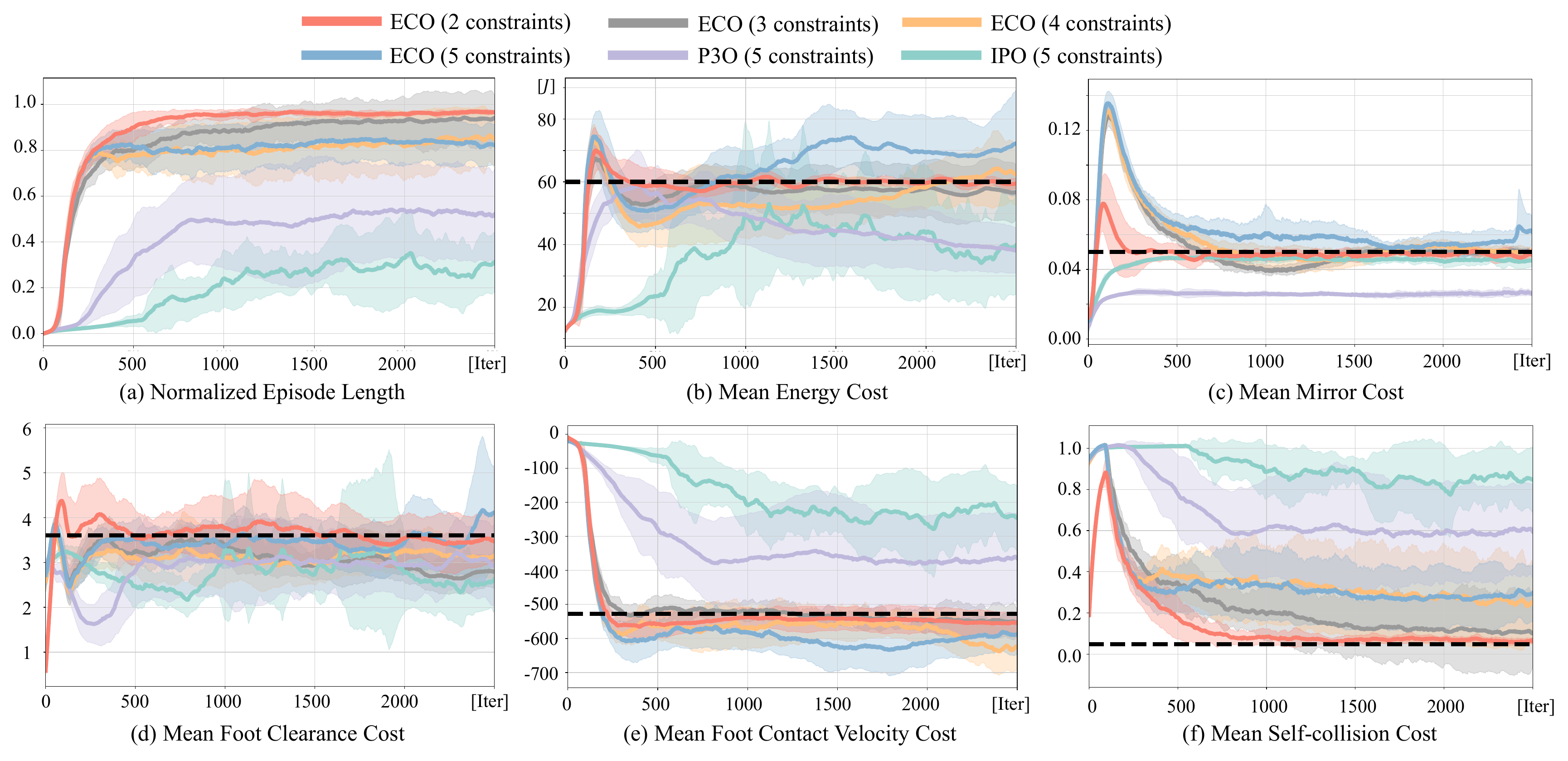}
\caption{\textbf{Training performance with more constraints settings}. Comparison of training metrics for ECO, P3O, and IPO. The thresholds for energy consumption, mirror reference motion, foot clearance, foot contact velocity, and self-collision are set at $60J$, 0.05, 3.6, -528, and 0.048, respectively, as indicated by the \textbf{black dashed lines} in (b)-(f). Results averaged over 10 random seeds.}
\label{fig:5constraints}
\end{figure*}

\subsubsection{\textbf{Hyperparameter Sensitivity Analysis}}
To demonstrate that the proposed ECO framework requires less human effort in hyperparameter tuning, we conduct a sensitivity analysis by training ECO on a \(0.1m/s\) linear velocity tracking task with an energy threshold of \(b_1 = 60J\). We fix the initial value of the Lagrange multiplier at 0 and vary the learning rate (\(\beta_1 = 0.001, 0.01, 0.1\)). As shown in \cref{fig:hyper}(a), which illustrates the mean energy cost over iterations, all three learning rates converge to a similar energy cost of approximately \(60J\) after an initial period of fluctuation. This pattern suggests that our Lagrangian approach ensures effective energy convergence across different learning rates. \cref{fig:hyper}(b) shows the evolution of the Lagrange multiplier over iterations for different learning rates. A smaller learning rate causes the Lagrange multiplier to change more slowly, resulting in higher energy consumption initially due to constraint violations As energy increases, the multiplier gradually rises, enabling faster convergence to the feasible region. Finally, \cref{fig:hyper}(c) illustrates that our method achieves stable convergence across the range of learning rates.

%\cref{fig:hyper}(b) shows the evolution of the Lagrange multiplier over iterations for different learning rates. Although the multiplier exhibits higher peaks with a learning rate of \(\beta_1 = 0.1\), it eventually stabilizes similarly to the lower learning rates (\(\beta_1 = 0.001\) and \(\beta_1 = 0.01\)). 

\subsubsection{\textbf{More Constraints Setting}}
\label{sec:5constraint}
In previous work on quadruped robots~\cite{kim2024not, lee2024exploring}, additional reward terms such as self-collision, foot contact velocity, and foot clearance were integrated into the constraints to reduce the need for extensive reward scaling. We also attempted to incorporate these three constraints. Specifically, we trained an additional multi-head cost critic to estimate the cost returns for self-collision, foot contact velocity, and foot clearance. For these additional constraints, we used the converged reward values obtained from ECO (with two constraints) as the constraint thresholds. We analyze the performance of several algorithms under five constraints and investigate how the convergence properties of ECO are affected by different constraint configurations.

For the setup with five constraints, ECO is trained with a fixed initial Lagrange multiplier \(\lambda_i = 0.0\) and a learning rate \(\beta_i = 1e-3\), while IPO and P3O used the same \(\kappa = 1.0\) for each constraint. As shown in \cref{fig:5constraints}(a), we observed that the episode lengths for ECO, IPO, and P3O with five constraints are significantly reduced compared to ECO with only two constraints. \cref{fig:5constraints}(e) shows that P3O and IPO fail to converge on the foot velocity contact constraint, while ECO demonstrates better constraint satisfaction. \cref{fig:5constraints}(f) shows that all algorithms fail to converge on the self-collision cost. This is possibly due to predicting self‐collision from the current privileged observation space is challenging. 

In the four-constraint setting of ECO (without self-collision), we observe moderately better satisfaction of the energy constraint compared to the five-constraint case. Further removing the foot contact velocity constraint (three constraints setting) led to even longer episode lengths than the four-constraint setting, although performance remained below that of the original two-constraint configuration. Given the empirical results and our previous observation of the one-constraint setting—where omitting the symmetric reference motion constraint led to unnatural walking behaviors, as demonstrated in the accompanying video—we conclude that the two-constraint configuration (energy consumption and symmetry constraints) is the most suitable for achieving energy-efficient and consistently stable humanoid walking in our task. Further discussion and rationale are provided in \cref{sec:discussion_multiconstraint}.

%In summary, we empirically find that using two constraints leads to energy-efficient and consistently stable humanoid locomotion. For further discussion, see \cref{sec:discussion_multiconstraint}.}

\begin{figure}[t!]
\centering
\includegraphics[width=0.85\linewidth,trim=0cm 4.2cm 21cm 0.0cm,clip]{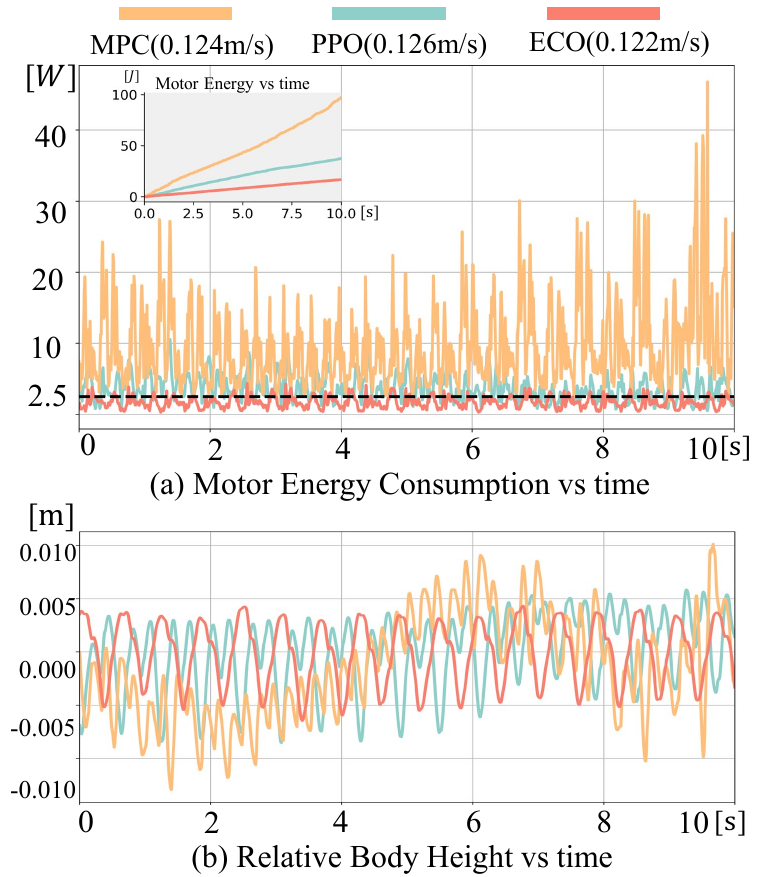}
\caption{\textbf{Motor energy consumption and body height oscillation in the real world.} Single-run deployment curves are shown for visual clarity. The relative body height is defined as the offset from the setpoint of the body height. The inset plot in (a) is the cumulative motor energy.}
\label{fig:real_data}
\end{figure}

\section{Experiments}\label{sec:exp}
\subsection{Real World Results}
\begin{figure}[t!]
\centering
\includegraphics[width=\linewidth,trim=0cm 10cm 10cm 0cm,clip]{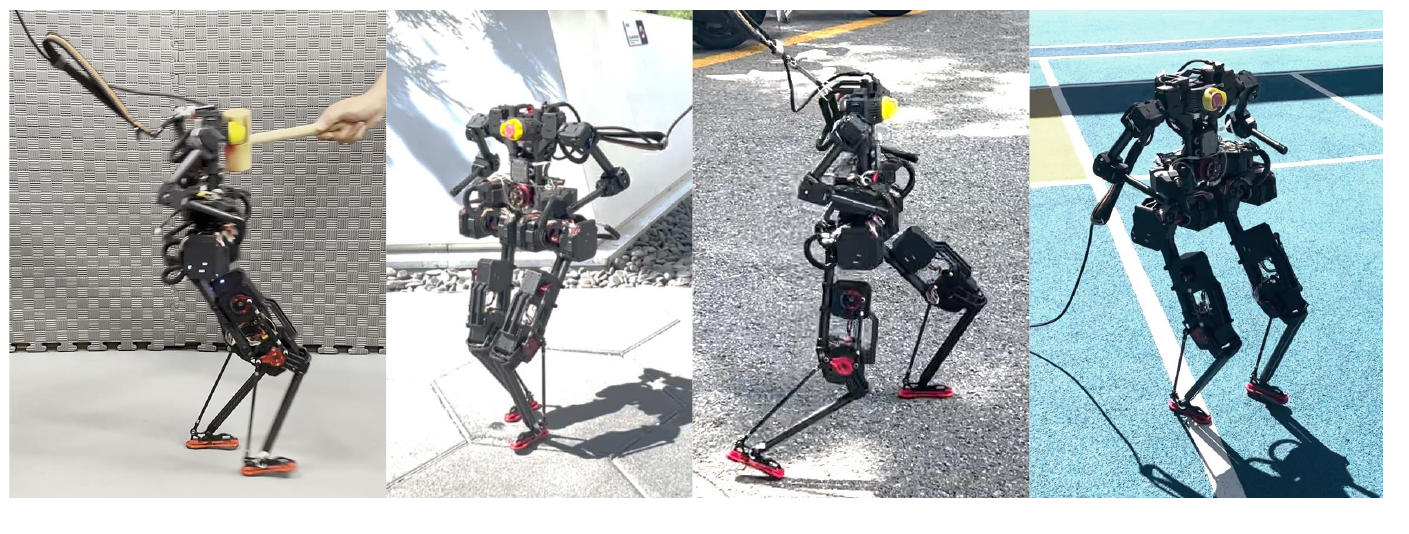}
\caption{\textbf{Screenshots of the sim-to-real transfer.} BRUCE successfully rejects disturbance and traverses outdoor terrains which can be seen in the accompanying video.}
\label{fig:real_wild}
\end{figure}

To verify the improvement in energy efficiency, we deploy the learned policy from ECO, along with the two baseline methods, on the real robot hardware BRUCE for comparison. Statistical results across multiple experiment runs, including the linear velocity along the x direction (speed), the total motor power (energy), and the body height in the world frame, are reported in \cref{tab:algo_performance_real}. Due to the sim-to-real gap, velocity tracking on the hardware is less accurate than in simulation. To ensure a fair comparison of energy consumption, we select three sets of data with similar, though not identical, speeds from multiple experiments using different methods. As shown in \cref{fig:real_data}(a), ECO consistently maintains motor power near or below the energy cost threshold $2.5W(60J/24s)$—approximately 6 times lower than MPC and 2.3 times lower than PPO according to the cumulative energy graph. Moreover, with the learned policy from ECO, BRUCE successfully rejects disturbances and traverses different outdoor terrains as illustrated in \cref{fig:real_wild}. The full hardware demonstration clips including comparisons with baselines, disturbance rejection, and outdoor walking can be seen in the accompanying video.

%Meanwhile, we compare the body oscillation over a $10s$ stable walking period. \cref{fig:real_data}(b) shows that BRUCE exhibits fewer oscillations with a lower frequency and smaller magnitude when walking with the policy from ECO, compared to MPC and PPO. This suggests that improved energy efficiency also results in lighter steps and reduced stomping. 
%\cref{fig:real_data2} further illustrates the joint-level behaviors during walking. \cref{fig:real_data2}(a) compares the left knee joint angle over time, where ECO demonstrates smoother and less flexed joint movement compared to MPC and PPO. A less flexed knee indicates improved energy efficiency by minimizing excessive joint movement during each step. In \cref{fig:real_data2}(b), the torque applied to the left knee joint is compared. ECO consistently requires lower torque to maintain stability and velocity tracking, resulting in reduced energy expenditure during walking. These observations align with the cumulative energy consumption results in \cref{fig:real_data}(a), confirming that ECO not only reduces overall energy consumption but also improves the mechanical efficiency of joint-level control in real-world walking scenarios.

\renewcommand{\arraystretch}{1.2}
\begin{table}[t!]
\caption{\textbf{Algorithm performance comparison in the real world.} Aggregated over 10 runs, each lasting 10 seconds.}
\centering
\resizebox{\linewidth}{!}{
\begin{tabular}{cccc}
\toprule
& Speed (m/s) & Energy (w)
& Body Height (m) \\
\hline
MPC 
& $0.124 \pm \mathbf{0.0072}$ & $9.1176 \pm 5.0029$ & $0.3483 \pm 0.0056$ \\
\hline
PPO 
& $0.126 \pm 0.0088$ & $3.9080 \pm 1.8193$ & $0.3745 \pm 0.0034$ \\
\hline
\rowcolor{Gray!15} \textbf{ECO}
& $0.122 \pm 0.0089$ & $\mathbf{1.7490} \pm \mathbf{0.8474}$ & $0.3790 \pm \mathbf{0.0030}$ \\
\bottomrule
\end{tabular}
}
\label{tab:algo_performance_real}
\end{table}

\subsection{Emergent Behaviors}
Empirical studies indicate that straight knee and heel-to-toe walking enhance energy efficiency in humans and humanoid robots~\cite{fasano2024efficient, adamczyk2006advantages}. Instead of manually designing these behaviors for efficient locomotion, ECO shows that minimizing energy consumption can generate similar behaviors without explicit prescriptions, which also demonstrates that these bio-inspired behaviors are related to energy efficiency.

\begin{figure}[t!]
\centering
\includegraphics[width=0.85\linewidth,trim=0.2cm 2.5cm 18.8cm 0.0cm,clip]{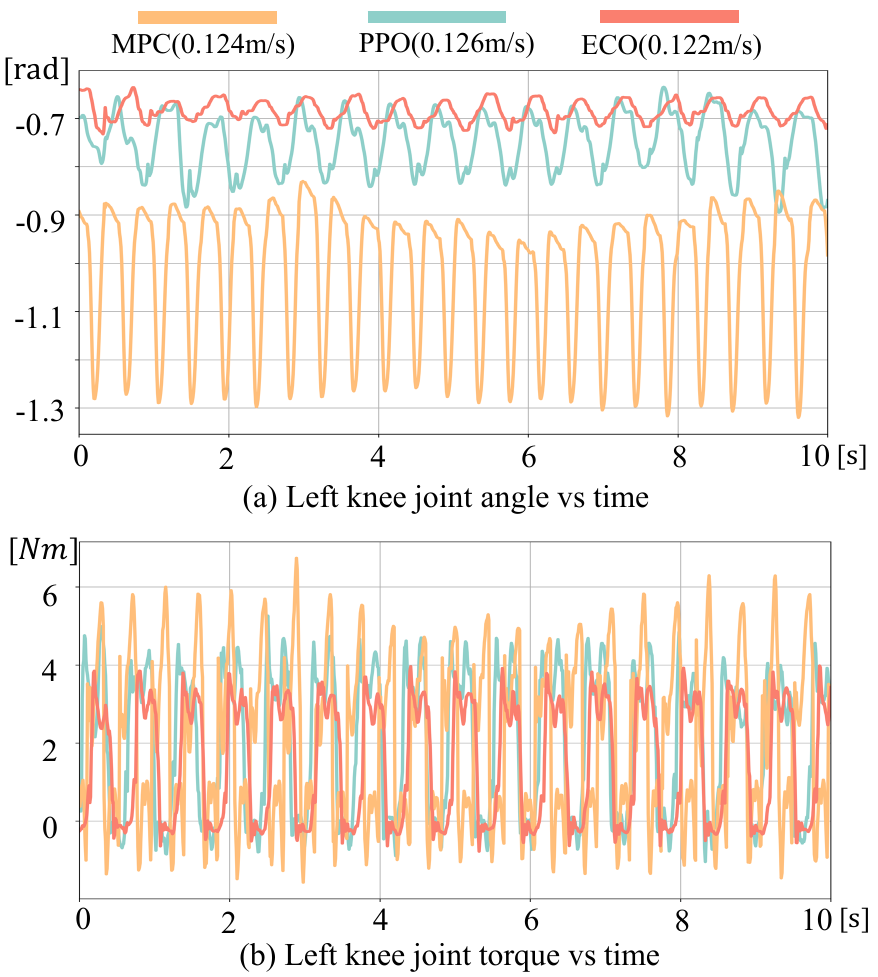}
\caption{\textbf{Comparison of left knee joint angle and torque between different methods in the real world.} We present single-run deployment curves for visual clarity. A larger knee joint angle reflects a more flexed knee during walking, while a smaller knee joint torque at ground contact indicates a lighter step.}
\label{fig:real_data2}
\end{figure}

\cref{fig:real_data2} illustrates the joint-level behaviors during walking. \cref{fig:real_data2}(a) compares the left knee joint angle over time, where ECO demonstrates a less flexed knee compared to MPC and PPO. In \cref{fig:real_data2}(b), the torque applied to the left knee joint is compared. ECO consistently requires lower torque to maintain stability and velocity tracking, resulting in reduced energy expenditure during walking. Due to the lack of precise force-torque sensors under the feet of BRUCE, the knee torques can also implicitly serve as a measure of the intensity of ground contact as the knee exerts the most force in humanoid locomotion. Additionally, we compare the body oscillation over a $10s$ stable walking period. \cref{fig:real_data}(b) shows that BRUCE exhibits fewer oscillations with a lower frequency and smaller magnitude when walking with the policy from ECO, compared to MPC and PPO. The less body oscillation and smaller knee torques suggest that improved energy efficiency also results in lighter steps and reduced stomping. 
% These observations align with the cumulative energy consumption results in \cref{fig:real_data}(a), confirming that ECO not only reduces overall energy consumption but also improves the mechanical efficiency of joint-level control in real-world walking scenarios.

However, a heel-to-toe transition is not presented in the resulting policy as shown in \cref{fig:real_wild}. We argue that the heel-to-toe strategy, given the current hardware design of BRUCE, demands additional control effort to execute effectively. Furthermore, it requires a higher swing leg lift, which increases energy consumption in the hip and knee motors.

\section{Discussion}\label{sec:discussion}

\subsection{Rewards v.s. Constraints}
\label{sec:discussion_multiconstraint}
In this study, we find that tuning reward scales through hyperparameter search within the \ac{rl} framework (\ac{ppo}) is an inefficient approach to reducing energy consumption in humanoid robots. This inefficiency arises because reward scales do not correspond linearly to physical quantities, demanding a delicate balance between meeting energy requirements, following task commands, and maintaining motion stability. In contrast, constrained \ac{rl} explores a space defined by explicit constraints, many of which have direct physical significance, such as energy consumption limits and joint restrictions. This makes tuning constraints more straightforward and effective compared to adjusting reward or loss weights.

In this work, we try to incorporate self-collision, foot contact velocity, and foot clearance as additional constraints. Although such constraints have been successfully applied to quadruped robots~\cite{kim2024not, lee2024exploring}, we find it challenging to transfer these approaches directly to humanoids. This difficulty arises not only from technical challenges associated with the heterogeneity of constraint signals but also from differences in morphology and balance dynamics. First, the cost terms associated with different constraints often differ in magnitude, which can lead to imbalanced gradient updates when training the multi-head cost critic~\cite{kim2024not}. As a result, costs with smaller magnitudes—such as the self-collision cost in \cref{fig:5constraints}(f)—tend to converge more slowly. This issue could be mitigated by applying appropriate normalization techniques~\cite{hafner2023mastering} to rescale cost signals to comparable magnitudes. Second, accurately predicting different types of costs may require heterogeneous information. For instance, inferring self-collision from the current privileged observation (i.e., joint angles and base-link kinematics) is particularly challenging, as it requires the network to implicitly reconstruct the geometry of the body in the world frame from joint-centric data. This suggests that a richer observation space—including world-frame information—or a specialized cost critic for self-collision may be necessary to improve both prediction accuracy and constraint satisfaction.

Furthermore, humanoid robots have much smaller support polygons than quadruped robots, especially in the single-support phase, where the support region further shrinks. This renders the space of feasible trajectories much smaller than quadruped robots. In this case, imposing a large amount of feasibility constraints in constrained \ac{rl} may complex the search space. We believe further study is required to handle a larger number of constraints in humanoid locomotion. For instance, incorporating model-based prior knowledge to refine the search space—such as using nullspace projections to prioritize control tasks, as commonly done in the control community~\cite{dietrich2015overview}—could help in selecting appropriate constraints and mitigating task conflicts. A comprehensive investigation of such approaches is beyond the scope of this work and is left for future research.

\subsection{Optimization Method Selection}
Our experiments show that among the four constrained RL algorithms, PPO-Lag is the most effective. CRPO, as a primal algorithm, does not introduce additional dual variables for optimization and involves fewer hyperparameters, making it theoretically simpler and easier to implement compared to primal-dual methods. However, it lacks the smooth transition between reward and constraint optimization objectives that primal-dual methods offer, making it difficult to achieve stable convergence in humanoid robots. P3O uses clamping functions on constraints to apply a linear cost gradient penalty until the constraints are met. However, this linear penalty makes balancing multiple constraints difficult, as each constraint has distinct physical significance and magnitude, requiring separate coefficients $\kappa_i^{\text{P3O}}$. Small coefficients lead to slow convergence, while large ones make the system overly conservative. IPO, as a method that theoretically also employs penalty functions, exhibits performance similar to P3O in our experiments. It also requires careful tuning of the coefficient \(\kappa_i^{\text{IPO}}\) to balance constraints effectively. 

In contrast, ECO leverages PPO-Lag to update both the Lagrange multipliers and the policy, allowing the Lagrange multipliers and their learning rates to adapt to the magnitude of each constraint. By utilizing the Adam optimizer to update these multipliers and learning rates, ECO effectively adapts to each constraint while smoothly transitioning the optimization objective between rewards and constraints. As a result, ECO achieves stable convergence in humanoid robots. 

However, ECO faces challenges in optimizing the five constraints considered in our study, highlighting the need for improved optimization techniques in future work. Leveraging the dynamics model of humanoid robots to improve the convergence of constrained \ac{rl} within a restricted feasible region could be a promising direction~\cite{levy2024learning, huang2023safe, as2022constrained}.

\subsection{Sim-to-Real Failure Cases}  
Our methods successfully transfer to the real world, and we identify several key factors that critically affect policy transfer:

\begin{enumerate}[leftmargin=1.5em]
    \item \textbf{PD Gain Tuning}: We find that carefully tuning the proportional ($K_p$) and derivative ($K_d$) gains of the low-level PD controller is essential to align the joint-position tracking performance between simulation and hardware. Otherwise, differences in system dynamics can cause the policy to produce divergent behaviors in the real world.
    
    \item \textbf{Compliant ankle strategy}: We find that high PD gains at the ankle lead to instability upon ground contact due to sensor noise. To address this, we reduce the ankle’s PD gains, making the joints more compliant and the policy less sensitive to noise. This simple adjustment significantly enhances robustness to terrain variations and modeling errors in real-world tests.
    
    \item \textbf{Sensor Noise and Bias}: Real sensors exhibit both high-frequency noise (especially in joint-velocity measurements) and low-frequency biases (e.g.\ Euler-angle drift upon reset).  To account for this, we inject noise and random biases—matched in magnitude to hardware measurements—into the corresponding state observations during domain randomization.
\end{enumerate}

\subsection{More complex settings}
While our method is effective for stable walking, applying it to more complex scenarios, such as transitioning from flat surfaces to stairs, presents new challenges. For instance, a lower foot lift is energy-efficient on flat surfaces, but stair climbing requires higher foot lifts and precise planning to navigate varying stair heights efficiently. One possible extension is to introduce mode-aware constraints~\cite{sleiman2023versatile, zhang2024learning} that adapt according to different terrains and command types. For example, energy constraints could be relaxed when stair climbing is anticipated. A terrain classifier or visual encoder~\cite{agarwal2023legged, chane2024soloparkour, yu2021visual} could be used to identify terrain types, allowing the policy to dynamically adjust constraint thresholds accordingly. In addition, model-based planning~\cite{hansen2024hierarchical, huang2023safe, ma2022combining} can be incorporated, in which a high-level planner selects locomotion modes based on the environment, while the low-level control executes constrained locomotion policy conditioned on the selected mode. This hierarchical approach aligns well with the constrained RL framework, as constraints can be defined and enforced differently across modes. We consider this an exciting future direction and plan to explore model-based extensions in our follow-up work.

% \subsection{Energy Efficiency v.s. Walking Robustness}
% We studied that within \ac{ppo} framework, including energy efficiency along with walking stability and velocity tracking lead to conflict objectives. The learned policy either sacrifice energy efficiency or walking robustness. On the contrary, within the proposed ECO framework, we verified that improving energy efficiency does not sacrifice walking robustness through extensive experiments, providing a new perspective to balance multiple objectives of \ac{rl} frameworks. 

\section{Conclusion} \label{sec:conclusion}
In this paper, we optimize the energy consumption for humanoid walking through constrained RL by separating energy consumption and reference motion objectives from the reward function. We demonstrate that the ECO framework can converge to a policy that achieves lower energy consumption while maintaining robustness. Meanwhile, ECO is resilient to hyperparameter settings, thereby eliminating the need for extensive tuning on the weights of rewards. Extensive experiments in both simulation and real-world hardware validate that our approach can significantly improve the energy efficiency of humanoid walking without compromising robustness.

{
%\balance
\bibliographystyle{ieeetr}
\bibliography{reference}

%\clearpage

\vskip -1\baselineskip plus -1fil
\begin{IEEEbiography}
[{\includegraphics[width=1in,height=1.25in,clip,keepaspectratio]{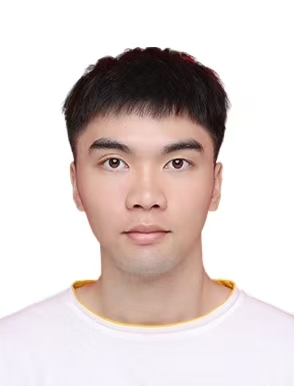}}]{Weidong Huang}
received the M.S. degree from the Beihang University in 2024. He is now a research engineer at State Key Laboratory of General Artificial Intelligence, Beijing Institute for General Artificial Intelligence (BIGAI). He received his B.S. degree in South China Normal University in 2021. His research interests include robotics, reinforcement learning, and model-based planning.
\end{IEEEbiography}

\vskip -1\baselineskip plus -1fil
\begin{IEEEbiography}
[{\includegraphics[width=1in,height=1.25in,clip,keepaspectratio]{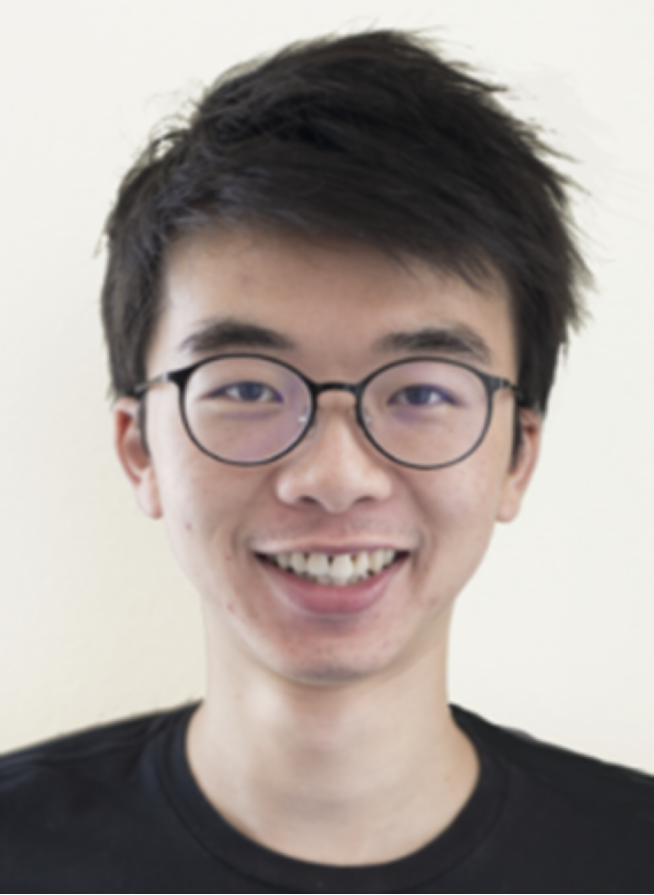}}]{Jingwen Zhang}
(Member, IEEE) received the B.S. degree and the M.S. degree from the Department of Mechanical Engineering of Tsinghua University and Columbia University in 2015 and 2017, and the Ph.D. degree from the Department of Mechanical and Aerospace Engineering, University of California, Los Angeles in 2023. He is now a research scientist at State Key Laboratory of General Artificial Intelligence, Beijing Institute for General Artificial Intelligence (BIGAI). His research interests include robotics, optimal control, and motion planning.
\end{IEEEbiography}

\vskip -1\baselineskip plus -1fil
\begin{IEEEbiography}[{\includegraphics[width=1in,height=1.25in,clip,keepaspectratio]{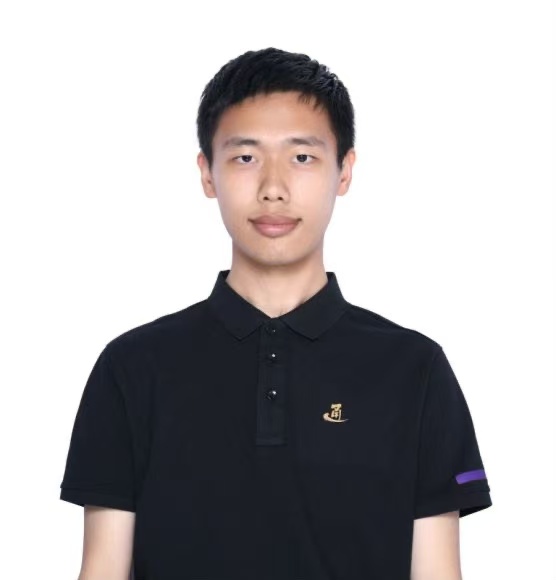}}]{Jiongye Li}
is now a third-year student in the Department of Automation of Tsinghua University. He is now a intern at State Key Laboratory of General Artificial Intelligence, Beijing Institute for General Artificial Intelligence (BIGAI). His research interests include humanoid robots, reinforcement learning and optimal control.
\end{IEEEbiography}

\vskip -1\baselineskip plus -1fil
\begin{IEEEbiography}[{\includegraphics[width=1in,height=1.25in,clip,keepaspectratio]{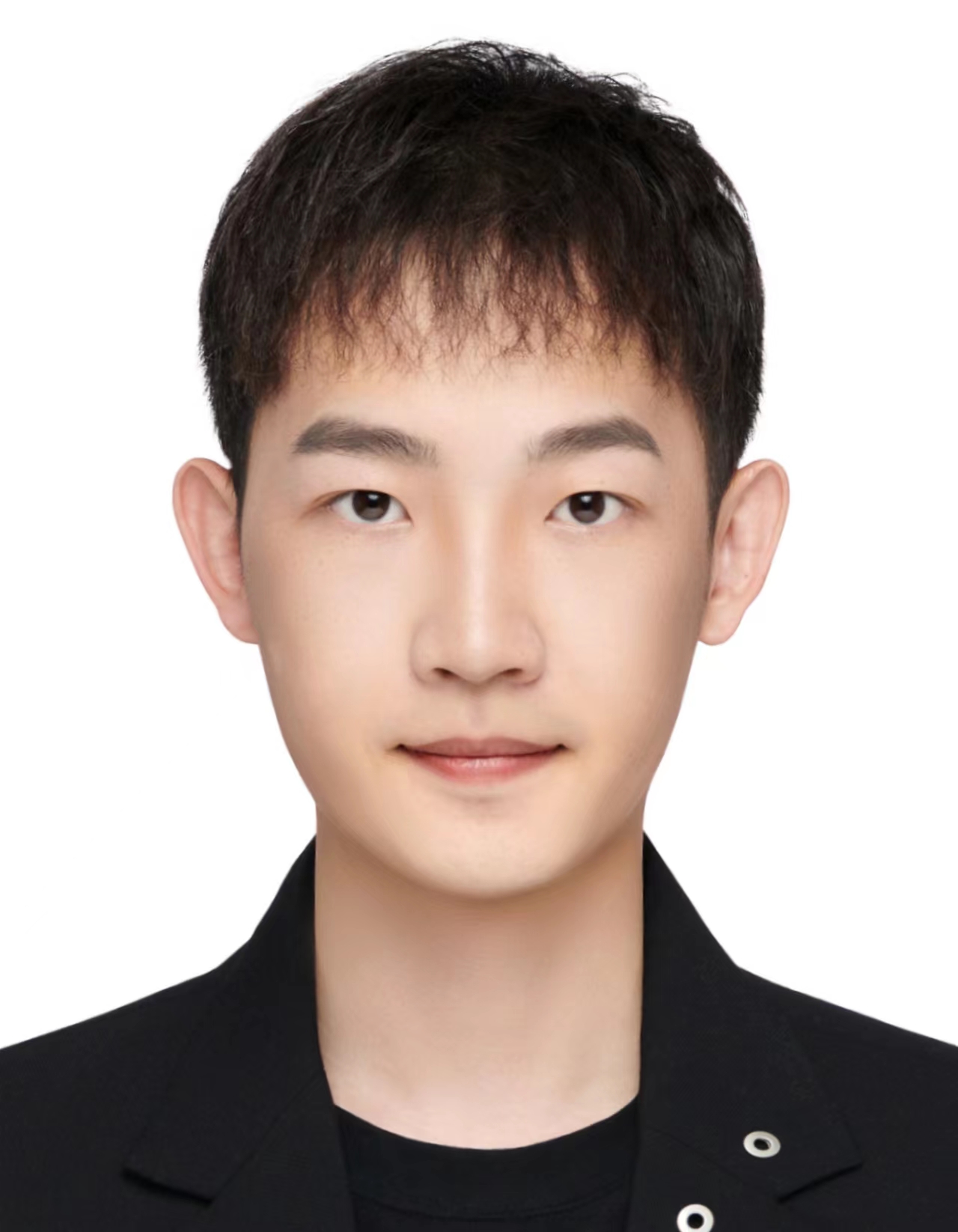}}]{Shibowen Zhang}
received the B.E. degree in Control Science and Engineering, from Tongji University, Shanghai, China, in 2024. He is currently a first-year Ph.D. student at the School of Control Science and Engineering, University of Science and Technology of China (USTC). His research interests include robotics and control.
\end{IEEEbiography}

\vskip -1\baselineskip plus -1fil
\begin{IEEEbiography}[{\includegraphics[width=1in,height=1.25in,clip,keepaspectratio]{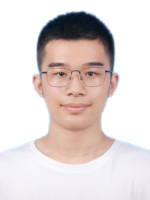}}]{Jiayang Wu}
is currently a senior undergraduate student in the Department of Computer Science and Technology at Harbin Institute of Technology (HIT). He is also an intern research assistant at the Beijing Institute for General Artificial Intelligence (BIGAI). His research interests include robotics, optimal control, and reinforcement learning.
\end{IEEEbiography}

% \vskip -1\baselineskip plus -1fil
% \begin{IEEEbiography}[{\includegraphics[width=1in,height=1.25in,clip,keepaspectratio]{authors/SCZhu}}]{Song-Chun Zhu}(Fellow, IEEE)
% received a Ph.D. degree from Harvard University in 1996, and is a Chair Professor jointly with Tsinghua University and Peking University, Dean of Institute for Artificial Intelligence at Peking University. He worked at Brown, Stanford, Ohio State, and UCLA before returning to China in 2020 to launch a non-profit organization---Beijing Institute for General Artificial Intelligence (BIGAI). He has published over 300 papers in computer vision, statistical modeling and learning, cognition, language, robotics, and AI. He received the Marr Prize in 2003, the Aggarwal prize from the Intl Association of Pattern Recognition in 2008, the Helmholtz Test-of-Time prize in 2013, twice Marr Prize honorary nominations in 1999 and 2007, the Sloan Fellowship, the US NSF Career Award, and the ONR Young Investigator Award in 2001. He serves as General co-Chair for CVPR 2012 and CVPR 2019. 
% \end{IEEEbiography}

\vskip -1\baselineskip plus -1fil
\begin{IEEEbiography}[{\includegraphics[width=1in,height=1.25in,clip,keepaspectratio]{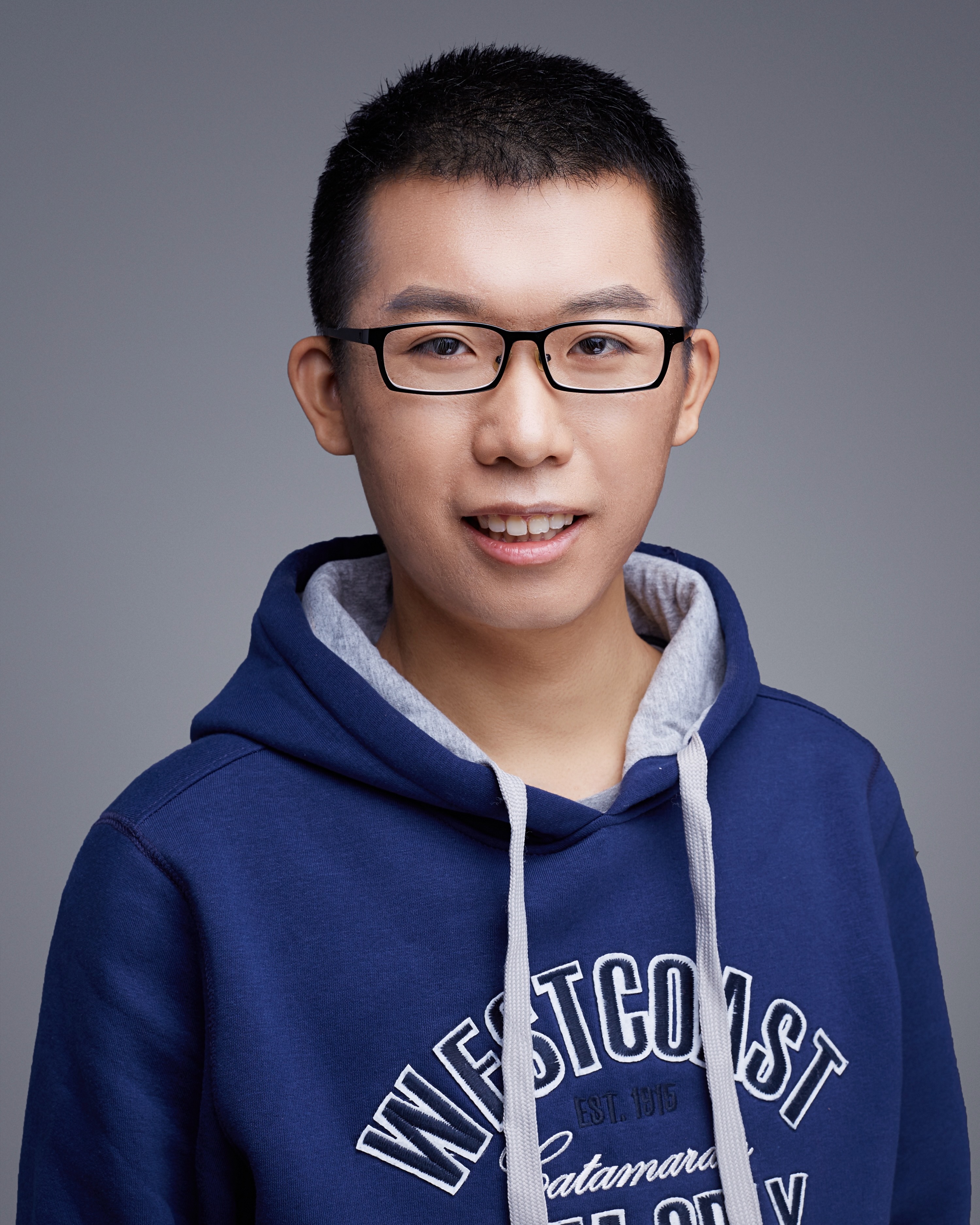}}]{Jiayi Wang}
(Member, IEEE) received his Ph.D. degree in robotics from the University of Edinburgh in 2023. He is currently a Research Scientist at Beijing Institute for General Artificial Intelligence (BIGAI). Before that, he was a research associate at the University of Edinburgh in 2024 and was a research intern at the Italian Institute of Technology in 2017. 
His research interests include humanoid locomotion, multi-contact motion planning, reinforcement learning, and trajectory optimization. 
\end{IEEEbiography}

\vskip -1\baselineskip plus -1fil
\begin{IEEEbiography}[{\includegraphics[width=1in,height=1.25in,clip,keepaspectratio]{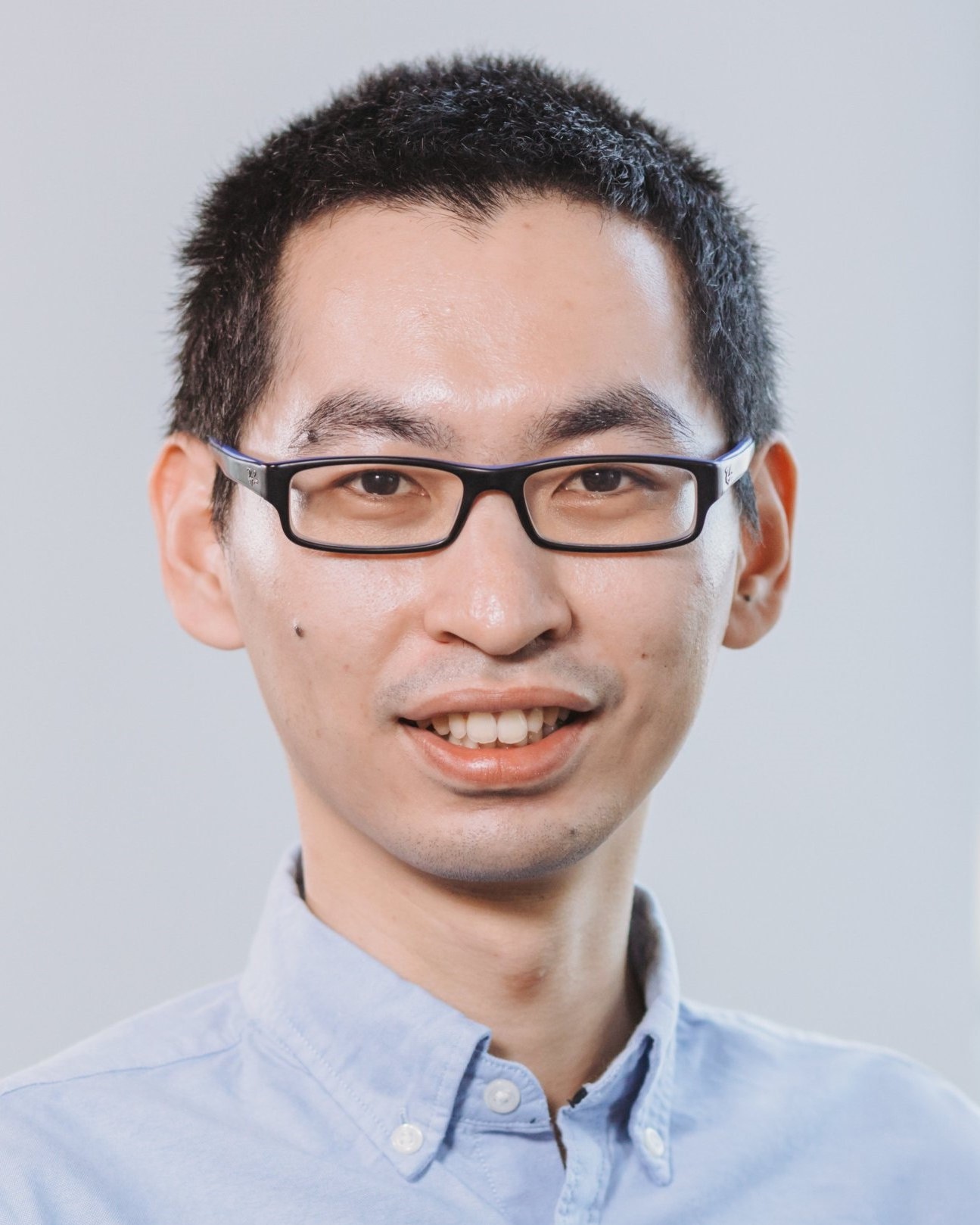}}]{Hangxin Liu}
(Member, IEEE) received his Ph.D. degree in Computer Science from the University of California, Los Angeles (UCLA) in 2021. He is currently a research scientist at State Key Laboratory of General Artificial Intelligence, Beijing Institute for General Artificial Intelligence (BIGAI). He received an M.S. degree in Mechanical Engineering from UCLA in 2018 and two B.S. degrees in Mechanical Engineering and Computer Science, both from Virginia Tech in 2016. His research interests focus on robot perception, learning, human-robot interaction, and cognitive robotics. 
\end{IEEEbiography}

\vskip -1\baselineskip plus -1fil
\begin{IEEEbiography}[{\includegraphics[width=1in,height=1.25in,clip]{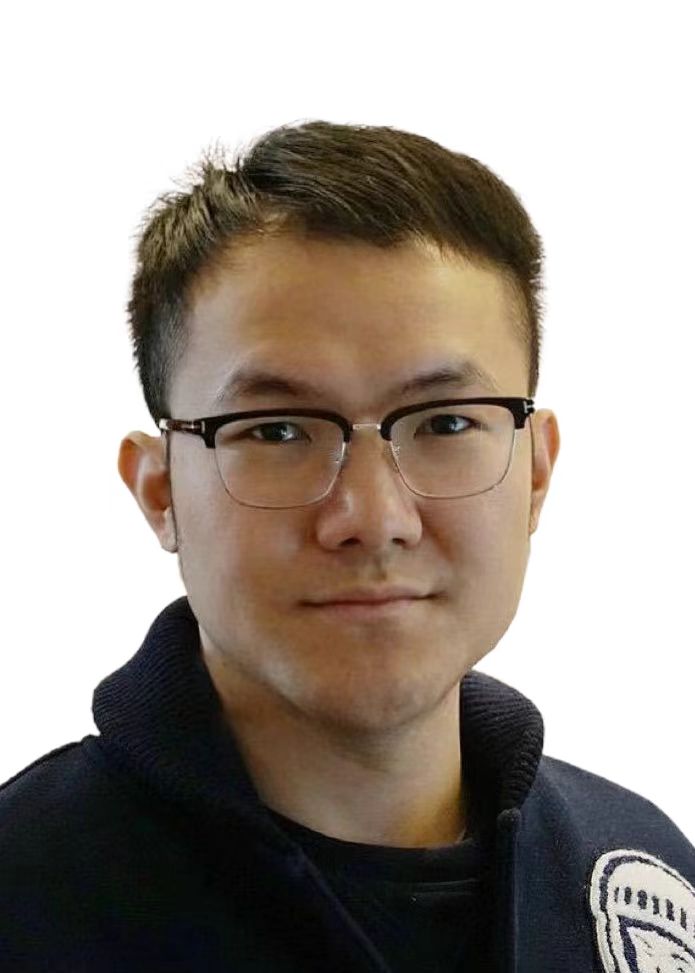}}] {Yaodong Yang} (Member, IEEE) received the bachelor’s degree from the University of Science and Technology of China, the MSc degree from Imperial College London, and the PhD degree from University College London (nominated by UCL for Joint AAAI/ACM SIGAI Doctoral Dissertation Award). He is an assistant professor with Institute for AI, Peking University. Before joining Peking University, he was an assistant professor with King’s College London. He studies game theory, reinforcement learning and multi-agent systems, aiming to achieve artificial general collective intelligence through multi-agent reinforcement learning. He has maintained a track record of more than sixty publications at top conferences (NeurIPS, ICML, ICLR, etc) and top journals (Artificial Intelligence, National Science Review, etc), along with the best system paper award at CoRL 2020 and the best blue-sky paper award at AAMAS 2021. He was awarded ACM SIGAI China Rising Star and World AI Conference (WAIC’22) Rising Star.
\end{IEEEbiography}

\vskip -1\baselineskip plus -1fil
\begin{IEEEbiography}[{\includegraphics[width=0.9in,height=1.25in,clip,keepaspectratio]{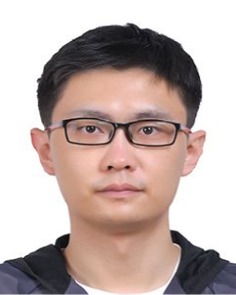}}]{Yao Su}
(Member, IEEE) received the B.S. degree from the School of Mechatronic Engineering, Harbin Institute of Technology in 2016, and the M.S. and Ph.D. degrees from the Department of Mechanical and Aerospace Engineering, University of California, Los Angeles (UCLA) in 2017 and 2021. He is now a research scientist at State Key Laboratory of General Artificial Intelligence, Beijing Institute for General Artificial Intelligence (BIGAI). His research interests include robotics, control, optimization, trajectory planning, and mechatronics.
\end{IEEEbiography}
}
\end{document}